\documentclass[11pt]{article}
\usepackage[T1]{fontenc}
\usepackage{lmodern}
\usepackage[utf8]{inputenc}
\usepackage[round,authoryear]{natbib}
\let\cite\citep
\usepackage{graphicx, float, placeins, booktabs, tabularx}
\usepackage{amssymb}
\usepackage{amsthm}
\usepackage{comment}
\usepackage[margin=1in]{geometry}
\usepackage{caption}
\usepackage{tikz}
\usepackage{amsmath}
\usepackage{booktabs}
\usepackage{enumitem}
\usepackage{adjustbox}

\usepackage{xstring}  

\newcounter{concept}
\newcommand{\currentconceptprefix}{} 

\newcommand{\concept}[3]{%

  \ifx\currentconceptprefix\empty
    \def\currentconceptprefix{#1}%
  \fi
  \IfStrEq{\currentconceptprefix}{#1}{}{%
    \setcounter{concept}{0}%
    \def\currentconceptprefix{#1}%
  }%

  \stepcounter{concept}%
  \expandafter\xdef\csname concepttitle@#3\endcsname{#1\arabic{concept}. #2}%
  \par\noindent\textbf{#1\arabic{concept}. #2}\label{#3}\par\vspace{0.5em}
}

\newcommand{\conceptref}[1]{\hyperref[#1]{\csname concepttitle@#1\endcsname}}

\usepackage[most]{tcolorbox}
\usepackage[hidelinks]{hyperref} 
\newtcolorbox{biasbox}[1]{%
  colback=gray!5,           
  colframe=gray!60!black,   
  fonttitle=\bfseries,      
  title={#1},               
  sharp corners,
  boxrule=0.5pt,
  arc=2mm,
  left=4mm,
  right=4mm,
  top=2mm,
  bottom=2mm,
  breakable
}

\theoremstyle{definition}

\begin{document}
\title{The Gold in Bias: Maturing the AI Design Process through Verification}
\author{Samira Maghool\thanks{Corresponding author. Email: \texttt{samira.maghool@unipegaso.it}. ORCID: 0000-0001-8310-2050.}\\
\small Pegaso University, Italy\\
\and
Paolo Ceravolo\thanks{Email: \texttt{paolo.ceravolo@unimi.it}. ORCID: 0000-0002-4519-0173.}\\
\small University of Milan, Italy}
\date{}
\maketitle

\begin{abstract}
{\bf Background:} 
Bias in AI systems is typically framed as a flaw to be minimized, yet it also serves as a critical indicator of underlying weaknesses in data, modeling assumptions, and system design. Existing approaches often treat bias as an isolated problem rather than as evidence that can strengthen verification and governance across the AI lifecycle.      
    
{\bf Objectives: } This paper aims to reconceptualize bias as a diagnostic tool that supports rigorous AI verification. We seek to develop a multidimensional framework to analyze bias, demonstrate how biases emerge in both Traditional and Generative AI, and provide a structured pathway for verification-driven mitigation.

{\bf Methods: } We present a multidimensional framework analyzing bias across four dimensions: origin sources, emergence points throughout the AI modeling lifecycle, technical and methodological causes, and validation approaches for detection and mitigation. Through a comprehensive typology spanning traditional and generative AI systems, we demonstrate how biases manifest and propagate across development stages. Our analysis encompasses $30$ distinct bias types, $16$ verification methods, and $20$ countermeasures, providing an actionable roadmap for practitioners. We introduce a hierarchical evidence framework that distinguishes internal validity (mechanistic integrity of AI systems) from external validity (contextual reliability in deployment environments).

{\bf Results: } The framework reveals how biases manifest and propagate across modeling stages, enabling systematic mapping between bias types, verification techniques, and effective countermeasures. The proposed evidence hierarchy clarifies how different verification strategies contribute to mechanistic integrity and contextual reliability.

{\bf Conclusions: } We advocate for ``Ethics by Design'' principles that integrate bias verification throughout the development lifecycle, enabling the construction of fairer, more robust, and trustworthy AI systems.
    
\end{abstract}

\section{Introduction}

The rapid adoption of Artificial Intelligence (AI) across industry, administration, and other domains requires \textbf{rigorous scrutiny} of its \textbf{design and evaluation} -- especially as AI systems increasingly influence critical decisions, from medical diagnoses to automated audit procedures~\cite{li2023trustworthy,wang2022business}. Historically, AI system design has emphasized the \textbf{technical correctness of learning procedures}, focusing on whether algorithms are properly implemented and trained on carefully curated \textit{ground truth} datasets~\cite{lakshmanan2020machine}. In this paradigm, ground truth is often considered the sole source of validity, with system performance judged almost entirely by its ability to reproduce or approximate labelled data. Although this approach has advanced optimization and benchmarking, it risks overlooking important factors such as data biases, misalignment with real-world conditions, limited model interpretability, and uneven performance across populations.

These considerations can be formalized through the lens of \textbf{internal} and \textbf{external validity}~\cite{mendling2025methodology}. \emph{Internal validity} refers to the extent to which observed model behavior can be attributed to intended design choices rather than confounding factors, including data representation, feature selection, and assumptions embedded in algorithmic reasoning. \emph{External validity} captures the degree to which model performance generalizes beyond controlled training and testing conditions, reflecting robustness and fairness across diverse populations, contexts, and applications. Both forms of validity are necessary for building AI systems that are not only technically correct but also trustworthy and socially reliable~\cite{wang2024rationality,panigutti2021fairlens,bansal2021most}.

Dysfunctional or biased outcomes in AI systems are increasingly recognized as \textbf{multifaceted}, resulting not only from flawed training data, but also from decisions throughout the design and deployment pipeline~\cite{shah2025comprehensive}. These problems often arise from the \textbf{cumulative effects} of decisions made at \textbf{different development stages}~\cite{prado2020bonseyes}. For example, algorithmic bias can result from apparently neutral preprocessing steps, such as image normalization techniques that inadvertently suppress features critical to certain groups~\cite{menezes2021bias}, or from problem formulation that overlooks social or cultural context~\cite{maghool2023enhancing}. Additionally, metrics emphasizing aggregate performance, such as overall accuracy, can obscure significant \textbf{differences across subgroups}~\cite{huang2021performance}. Defining such subgroups is critical, as it determines which forms of variation become visible~\cite{10.1007/978-3-031-46846-9_1}. However, as illustrated by Simpson's paradox, aggregate-level performance gains can alter or reverse subgroup-level conclusions, making data segmentation choices determinant of assessment results~\cite{sharma2022detecting}.

Nevertheless, \textbf{bias} in AI systems is \textbf{not} an intrinsic or universally \textbf{objective property} -- it becomes meaningful only relative to defined goals or references~\cite{binns2018fairness,bueter2022bias}. What counts as ``biased'' depends on what the system is designed to optimize, who it serves, and what values are prioritized~\cite{mehrabi2021survey}. Identifying bias is not purely technical, but requires deliberate choices about performance benchmarks, fairness criteria, and social references~\cite{selbst2019fairness}. Moreover, verifying bias can involve different evidence levels, from statistical measures and technical checks to stakeholder feedback or real-world observations~\cite{bengio2024managing}.

A growing body of research has addressed AI bias from technical~\cite{de2023systematic,shah2025comprehensive}, ethical~\cite{binns2018fairness,bueter2022bias}, and social perspectives~\cite{soprano2024cognitive,selbst2019fairness}, ranging from algorithmic fairness (e.g., demographic parity, equalized odds)~\cite{mehrabi2021survey} to accountability mechanisms~\cite{wieringa2020account}, institutional governance~\cite{mittelstadt2019principles,messeri2024artificial}, and participatory design practices~\cite{spinde2025enhancing,katell2020toward}. However, bias need not be understood \textbf{solely as a flaw to be corrected}; it can also serve for critically examining and improving design practices. Indeed, it can serve as a \textbf{diagnostic lens}, an opportunity to uncover hidden assumptions, clarify value trade-offs, contextual constraints, and \textbf{strengthen design practices}. From this perspective, bias becomes a design tool for reinforcing the integrity, robustness, and \textbf{trustworthiness of AI systems}.

Embracing this perspective, our paper aims to illustrate \textbf{typical cases of bias} throughout the AI modelling lifecycle. Our objective is to provide a comprehensive view by explicitly linking each case to a specific bias type, the stages at which it emerges, the technical mechanisms causing it, and the validation instruments that can identify it.

To achieve this objective, we analyse bias across\textbf{ four distinct dimensions}. Each dimension introduces concepts that help characterise the different cases discussed in the paper and identify corresponding technical methods and practices.
More specifically, following the Introduction, Section~\ref{sec:AI} presents our conceptualization of bias and introduces the dimensions used to describe related concepts in the following subsections. Subsection~\ref{sec:origin} discusses the different sources of bias in AI systems, resulting in four main categories of distortion. Subsection~\ref{sec:lifecycle} outlines the main steps of the AI modeling lifecycle. Subsection~\ref{sec:model} provides an overview of technical flaws that may lead to bias in AI models, followed by Subsection~\ref{sec:evidence} that discusses the forms of evidence employed across validation approaches. 
Section~\ref{sec:exmp} illustrates examples categorized by the concepts from previous sections. Section~\ref{sec:verif} discusses different verification approaches. Section~\ref{sec:mitigation} presents mitigation strategies and countermeasures, while, Section~\ref{sec:conc} concludes with final remarks and directions for future research.

\section{Aim and Scope of the Study}\label{sec:aim}

This study aims to develop a unified framework that offers a \textbf{comprehensive and structured account} of the types of bias discussed in the literature. The final goal is to support developers, researchers, and regulators in recognising, analysing, and mitigating these biases, while positioning bias verification as a core element of \textbf{trustworthy, evidence-based AI design}. By systematically examining bias, we can identify recurring patterns that facilitate detection, as well as specific differences that necessitate targeted mitigation strategies.

Within this broader perspective, the paper considers bias across both \textit{Traditional AI (TAI)} and \textit{Generative AI (GenAI)}, treating it as a \textbf{context-dependent phenomenon} that can result in allocative harms in decision-oriented systems~\cite{raji2022actionable,norori2021addressing} and representational harms in content-generating models~\cite{watson2024algorithmic,birhane2023into}. We describe the distinct mechanisms through which these distortions arise and analyse how bias can emerge at any stage of the AI lifecycle, from problem formulation to deployment, highlighting the importance of continous verification mechanisms and effective countermeasures.

To address these challenges, we examine bias along four complementary dimensions:

\begin{enumerate}
    \item \textbf{A — Source of Bias:} conceptual origins and mechanisms that generate distortions relative to expected behaviour (Section \ref{sec:origin}).
    \item \textbf{B — AI Modeling Lifecycle:} stages at which distortions emerge, propagate, or intensify (Section \ref{sec:lifecycle}).
    \item \textbf{C — Technical and Methodological Limitations:} design or implementation issues that foster bias (Section \ref{sec:model}).
    \item \textbf{D — Validation Approaches:} methods for verifying and quantifying bias, organised into levels of evidence reflecting increasing rigour (Section \ref{sec:evidence}).
\end{enumerate}

\section{Methodological Approach}\label{sec:method}

This paper adopts a \textbf{design science research} approach, in which the objective is to develop a conceptual and methodological artefact that supports the understanding and diagnosis of bias across diverse AI systems. Our approach involves three interrelated phases that progressively deepen our understanding and refine the proposed framework.

\subsection{Phase 1: Conceptual Grounding and Structuring}
The conceptual grounding is formalised into four analytical dimensions (A--D) presented in Section \ref{sec:AI}, each composed of labelled concepts that are used consistently throughout the paper. These dimensions were developed through systematic literature review and synthesis of existing frameworks in AI bias research, fairness in machine learning, and AI verification methodologies. This structured taxonomy enables the precise classification of bias and establishes a shared vocabulary that links underlying mechanisms, lifecycle stages, and verification methods, thereby supporting a more systematic analysis of bias in AI systems.

\subsection{Phase 2: Cross-Dimensional Mapping of Bias Cases}
In the second phase, the paper conducts a systematic 
\textbf{cross-mapping of bias cases} using the four analytical dimensions. 

In Section \ref{sec:exmp}, 
30 examples relevant to both TAI and GenAI were collected and analysed. 
Examples were selected based on the following criteria: (i) documented evidence in peer-reviewed literature or authoritative technical reports; (ii) representation of diverse application domains (healthcare, finance, criminal justice, content generation); (iii) coverage of both traditional AI (TAI) and generative AI (GenAI) systems; (iv) illustration of distinct bias manifestation patterns across the AI lifecycle; and (v) availability of sufficient technical detail to enable cross-dimensional analysis.

To enhance readability and facilitate cross-references, each dimension is assigned a letter and each concept within a dimension is assigned a number. For instance, when we refer to \textbf{C2. Missing or incomplete data} in the text, we unambiguously indicate the second concept introduced in Section~\ref{sec:model}.
The mapping we propose clarifies the nature of the different examples we analyse, illustrating how complex distortions can be broken down into identifiable components. This decomposition facilitates both verification and mitigation strategies by revealing the specific mechanisms through which bias emerges and propagates.

\subsection{Phase 3: Integration into Evidence-Based Verification Frameworks}
The final phase synthesises the insights from the previous steps into 
\textbf{hierarchies of evidence} for internal and external validity presented in Section \ref{sec:verif}.  
These hierarchies represent different levels of rigour, from basic metric-based validation to approaches designed for continuous monitoring in production environments.
The resulting $16$ verification methods and $20$ countermeasures account for the differing nature of bias in TAI and GenAI systems, offering structured guidance for selecting appropriate evaluation strategies based on system type, risk level, and deployment context. Each verification method is mapped to the types of bias it can detect (Phase 2 taxonomy) and the forms of evidence it produces (internal vs. external validity).

Together, these three phases produce a coherent methodological artefact that connects 
conceptual definitions, lifecycle reasoning, technical analysis, and evidence-based 
verification, ultimately supporting the implementation of ``Ethics by Design'' principles in AI development.

\section{Conceptually Grounding Bias in AI}\label{sec:AI}

Bias represents a \textbf{distortion} that can compromise decision-making or goals such as accuracy, fairness, or robustness. Yet defining what counts as distortion is far from straightforward~\cite{bueter2022bias,murikah2024bias}. Bias is often seen as deviation from an objective truth, but this assumes rigid criteria. In practice, AI systems operate in dynamic environments where labels are subjective, data shift, and notions of correctness or fairness vary by context. An alternative interpretation views bias as deviation from a reference standard, such as data distribution, domain-specific validation practices, or normative expectations. In this view, bias becomes a \textbf{context-sensitive phenomenon}: what appears biased in one setting may reflect intentional alignment in another~\cite{binns2018fairness}.

This contextual nature becomes evident when design choices shape downstream outcomes~\cite{10.1613/jair.1.13196}. A healthcare triage model trained on biased data may deprioritize marginalized groups, while predictive policing can reinforce systemic inequalities. Similar techniques, however, raise fewer concerns in non-sensitive domains such as retail forecasting. These examples demonstrate that bias arises from the interplay of technical objectives, social values and application goals. As bias can arise at any stage of the AI lifecycle, it should be viewed as an opportunity to improve the quality of the design.

In this work we analyze bias in terms of its conceptual source (A) the lifecycle stage(s) at which it emerges (B), the technical or methodological limitations that are responsible for it (C), and the forms of evidence appropriate for detecting and interpreting it (D).

\subsection{Sources of Bias}\label{sec:origin}

To provide a structured overview of bias in AI, we begin by examining its \textit{general sources}, grounding our analysis in the categories of bias established in pre-AI literature that emphasize societal and human origins. This foundation is essential for two reasons. First, it connects our framework to existing scholarship on bias. Second, it enables us to trace how social context and human judgment influence the technical manifestations of bias in AI systems. Identifying the root sources before examining their technical instantiation allows us to better understand how societal prejudices, human subjectivity, and design choices shape AI outcomes across the development pipeline.
We categorize bias sources into three groups: Societal Prejudices, Human Subjectivity, and Technical Limitations.\\

\concept{A}{Societal prejudices}{sou:socpre}
This concept refer to systemic inequalities and social norms that may embed into AI systems. When training data or institutional practices reflect historical injustices, stereotypes or structural discrimination, AI systems reproduce and amplify these distortions. This may result in \textbf{Social Bias} in various forms such as \textit{Historical} and  \textit{Confirmation} bias which undermine fairness in areas such as hiring, lending, law enforcement, and healthcare~\cite{barocas2023fairness}. Rather than providing impartial decision-making, these systems risk exacerbating existing disparities.\\

\concept{A}{Human subjectivity}{sou:humsub} Human subjectivity introduces bias through individual judgment, selection, and cognitive constraints. Those involved in the design process, such as data annotators, developers, and decision-makers, bring their own implicit biases, cultural perspectives, and selective attention, which can lead to inconsistent or skewed inputs. This source of bias can cause \textbf{Data Representation Bias} once the data does not meet the data coverage requirements. Human subjectivity could also cause \textbf{Measurement and Decision Bias} in cases like biased threshold-setting~\cite{mitchell2019model} and \textbf{Usage Bias} when subjectivity leads to the system being misused. 
\\

\concept{A}{Technical limitations}{sou:teclim} Technical limitations generate bias through algorithmic constraints, imbalanced datasets, and narrow design choices that fail to generalise across populations. These produce \textbf{Data Representation Bias}, \textbf{Measurement and Decision Bias}, and \textbf{Usage Bias} when opaque interfaces hinder responsible application.

\begin{figure}[!htbp]
    \centering
    \includegraphics[width=\linewidth]{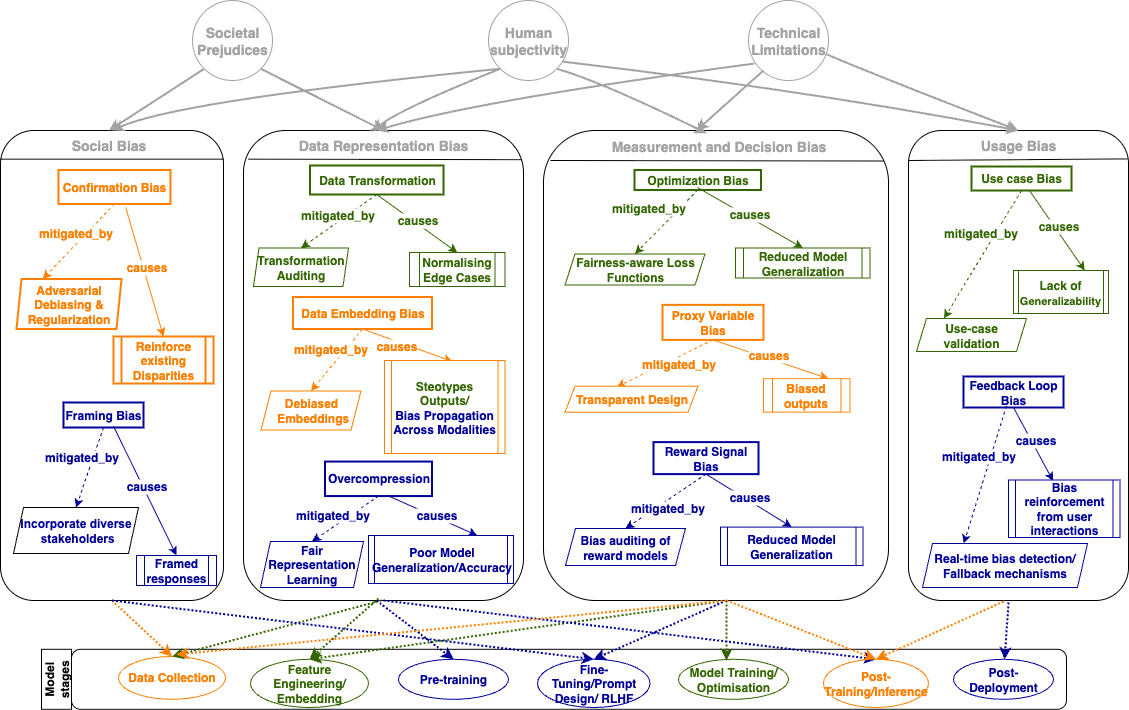}
    \caption{Categorization of different types of bias, Social Bias, Data Representation Bias, Measurement and Decision Bias, and Usage Bias, by their emergent sources: Societal Prejudices, Human Subjectivity, and Technical Limitation. For each category, a few examples of bias types are recalled. Framed blue examples demonstrate those types that, with high probability, happen in the GenAI system, while orange frames happen in both systems, and Green frames only for TAI. }
    \label{fig:bias_mixed}
\end{figure}

Figure~\ref{fig:bias_mixed} categorises bias types—Social Bias, Data Representation Bias, Measurement and Decision Bias, and Usage Bias—by their emergent sources: Societal Prejudices, Human Subjectivity, and Technical Limitations. \textcolor{blue}{Blue frames} indicate biases highly probable in GenAI, \textcolor{orange}{orange frames} denote biases in both GenAI and TAI, and \textcolor{green}{green frames} mark traditional AI-specific biases.

While valuable for tracing origins, this taxonomy alone does not guide intervention design, as biases emerge at different abstraction levels requiring stage-specific responses. We therefore supplement it by mapping each bias type to its corresponding lifecycle stage (Section \ref{sec:lifecycle}), technical limitations fostering its emergence (Section \ref{sec:model}), and appropriate validation approaches (Section \ref{sec:evidence}). This extended framework connects bias roots to concrete design and evaluation practices.

\subsection{The AI Modeling Lifecycle}\label{sec:lifecycle}

In order to understand where bias can emerge during the development of a system, it is helpful to outline the typical stages of AI modelling. While different works describe the pipeline with slight variations~\cite{gonzalez2023review}, most of the literature identifies the following stages:  \\

\FloatBarrier

\begin{table}[htbp]
\scriptsize
\centering
\caption{Stages of the AI lifecycle.}
\label{tab:AI_LC}
\begin{tabularx}{\textwidth}{|p{4cm}|X|}
\hline
\textbf{AI Modeling Stage} & \textbf{Description} \\ \hline

\concept{B}{Data Collection}{life:datcol} &
Gathering raw data from sensors, databases, user interactions, web scraping, or third-party providers. 
In \textit{Traditional AI} (TAI), data often include tabular formats, images, or domain-specific corpora. 
In \textit{Generative AI} (GenAI), large-scale multimodal datasets (text, images, audio, etc.) are collected, often from Internet-scale sources. \\ \hline

\concept{B}{Feature Engineering / Embedding}{life:feaeng} &
Transforms raw data into structured features or dense representations. 
In TAI, this includes manual feature selection (normalization, one-hot encoding, domain-specific extraction). 
In GenAI, it relies on automated embedding methods (tokenization, word/sentence or multimodal encoders). \\ \hline

\concept{B}{Pre-Training}{life:pretra} &
A defining step in GenAI, where foundational models (e.g., LLMs like GPT, vision models like CLIP) are trained on massive unlabeled or weakly labeled datasets to learn general-purpose representations. 
In TAI, this stage is often absent, as models are trained directly on task-specific data. \\ \hline

\concept{B}{Fine-Tuning / Prompt Design / RLHF}{life:finepro} &
Adapts or aligns generic models to specific tasks. Includes: 
\textbf{Fine-Tuning} — further training on smaller, task-specific datasets; 
\textbf{Prompt Design} — crafting inputs to steer outputs without modifying model parameters; 
\textbf{RLHF} — using human feedback to train a reward model that guides reinforcement learning for more aligned outputs. \\ \hline

\concept{B}{Model Training / Optimization}{life:modtra} &
In TAI, objectives focus on accuracy, precision, or recall, optimizing parameters to minimize loss. 
In GenAI, objectives include likelihood maximization, adversarial (GAN) losses, or transformer-specific training criteria. \\ \hline

\concept{B}{Deployment}{life:dep} &
The trained model is integrated into real-world environments where it performs or automates decisions. \\ \hline

\concept{B}{Post-Deployment}{life:posdep} &
Covers both inference and maintenance. 
In TAI, it produces predictions or scores; in GenAI, it generates new outputs (e.g., text, images). 
It includes decoding strategies (greedy, beam search, nucleus sampling), output calibration, and ongoing monitoring, retraining, auditing, and compliance updates. \\ \hline

\end{tabularx}
\end{table}

In the TAI lifecycle, \textbf{Data Collection}, \textbf{Feature Engineering/Embedding} are followed by \textbf{Model Training/Optimization} and end up with \textbf{Deployment} and \textbf{Post-Deployment} stages. While the GenAI systems' full stage implementation contains \textbf{Data Collection}, \textbf{Feature Engineering/Embedding}, \textbf{Pre-Training}, \textbf{Fine-Tuning/Prompt Design/RLHF}, \textbf{Inference}, \textbf{Deployment} and, \textbf{Post-Deployment}.

Figure~\ref{fig:bias_mixed} schematically outlines our approach in studying bias. In this approach, we move from Source to Mechanism (How it happens) to Manifestation (The resulting type of bias) and finally to the emergent Stage of the AI life cycle.



Although there is considerable overlap between different types of biases in TAI and GenAI models, the stages of the models prone to bias could differ.

\subsection{Bias in the Modeling Process}\label{sec:model}


Traditional evaluation exploits dataset \emph{ground truth} to compute quality metrics such as \textit{F1 score} or \textit{Area under the ROC curve} (AUC)~\cite{huang2021performance}. While quantitatively clear, these measures assume labels reliably represent objective truth, an assumption often violated when data fails to capture the full range of phenomena the deployed AI system will encounter.
Consequently, evaluation must move beyond model-data comparison to consider the broader \textit{target system} in which the model operates. Although the concept of the target system is abstract and cannot be denoted completely, it allows us to identify recurring flaws that can emerge throughout the design and development process.

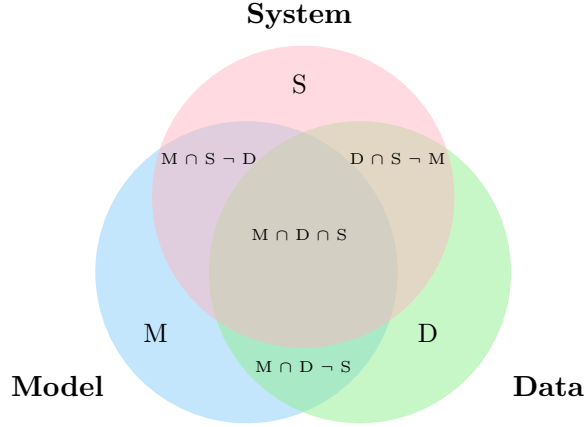
\begin{figure}[h!]
    \centering
\begin{tikzpicture}
  \definecolor{model}{RGB}{135,206,250}   
  \definecolor{data}{RGB}{144,238,144}    
  \definecolor{system}{RGB}{255,182,193}  

  \begin{scope}[opacity=0.5]
    \fill[model] (0,0) circle (2cm);
    \fill[data] (1.5,0) circle (2cm);
    \fill[system] (0.75,1) circle (2cm);
  \end{scope}

  \node at (-2.5,-1.5) {\textbf{Model}};
  \node at (4,-1.5) {\textbf{Data}};
  \node at (0.7,3.4) {\textbf{System}};
  \node at (-1.2,-0.8) {\small{M}};
  \node at (0.7,2.5) {\small{S}};
   \node at (2.4,-0.8) {\small{D}};

  \node at (0.75,-1.25) {\tiny M $\cap$ D $\neg$ S};
  \node at (2,1.5) {\tiny D $\cap$ S $\neg$ M};
  \node at (-0.5,1.5) {\tiny M $\cap$ S $\neg$ D};
  \node at (0.7,0.5) {\tiny M $\cap$ D $\cap$ S};

\end{tikzpicture}
\caption{This Venn diagram illustrates the three perspectives in model evaluation: M (what the model has learnt), D (what is present in the training and test data) and S (what is truly required by the system in the real world). Classical evaluation focuses on the intersection of M and D. Gaps between D and S or M and S, however, highlight areas where the model may fail to generalise or overfit to irrelevant patterns.}

    \label{fig:MDS}
\end{figure}

To frame our discussion, we can refer to the Venn diagram in Figure~\ref{fig:MDS}. This diagram illustrates the three crucial, and often distinct, domains of model evaluation: what the model has learned ($M$), the information encapsulated in the training and test data ($D$), and the true requirements of the real-world system ($S$).
Conventional evaluation methodologies primarily assess the intersection of $M$
and $D$ —that is, how well the model captures the statistical regularities encoded in its dataset. While this focus is necessary for internal validation, it can also be myopic because it overlooks other sources of failure.


Building on these conceptual distinctions, we can identify common aspects of AI modeling that lead to bias. The following subsections introduce seven common sources of bias in modeling. \\

\concept{C}{Biased data.}{mod:biadat} When the patterns captured by the model from the data do not authentically exist within the system, we are in the region $M \cap D \neg S$. In this situation, the model inherits distortions originating from the sources used to acquire the data. Such distortions often reflect stereotypes, sampling imbalances, or historical inequities embedded in the social, cultural, or institutional contexts in which the data were generated (Sec.~\ref{calt}). They may also emerge from temporal or contextual shifts, commonly referred to as \textit{concept drift}, where the data distribution evolves over time and no longer represents the current state of the system (Sec.~\ref{temp}). \\

\concept{C}{Missing or incomplete data.}{mod:misdat} 
When the available data provides only a partial representation of the system, the model is trained on an incomplete view of the phenomena it is meant to capture. This situation corresponds to the region $S \neg D \neg M$, where parts of the system are not reflected in the data. This type of incompleteness can be caused by limitations in the way data is collected, such as the omission of relevant cases, attributes, or subpopulations (such as Data Coverage Bias in Sec.~\ref{cov}). It can also result from measurement constraints, privacy restrictions or selective recording practices that exclude certain events or groups systematically (Sec.~\ref{meas}). Missingness becomes a source of bias when certain classes are not represented, or when the absence of data correlates with sensitive or outcome-related variables. This distorts the relationships that the model learns from the available evidence. \\

\begin{figure}[htbp]
    \centering
    \includegraphics[width=0.8\textwidth]{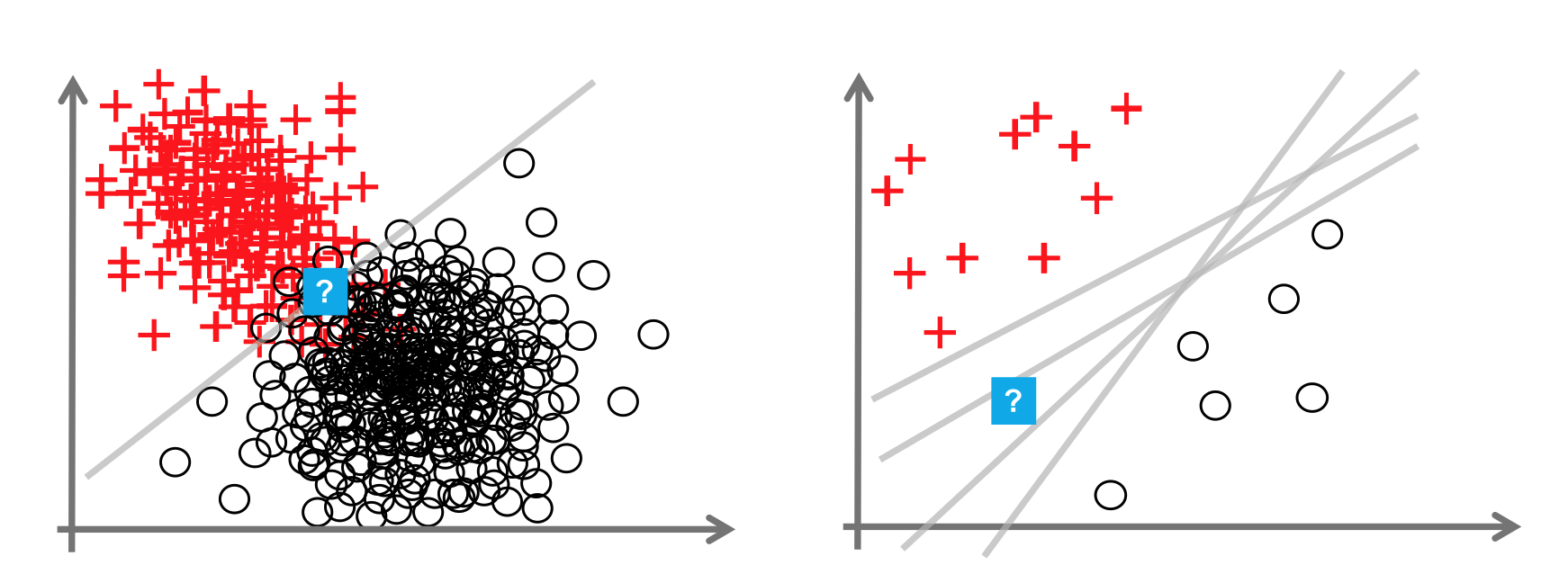} 
    \caption{Taken from~\cite{hullermeier2021aleatoric}. Illustration of systematic misclassification within the region $M \cap D \cap S$. The left panel depicts an \textit{aleatoric} condition, where class boundaries overlap and instances near the frontier are inherently ambiguous, leading to unavoidable misclassifications. The right panel shows an \textit{epistemic} condition, where instances occur in regions of the decision space that are underrepresented in the training data, resulting in uncertain or systematically incorrect predictions.} 
    \label{fig:uncertML} 
\end{figure}

\concept{C}{Systematic misclassification.}{mode:sysmis}
Even when model, data, and system overlap ($M \cap D \cap S$), bias may persist as systematic misclassification. As Fig.~\ref{fig:uncertML} illustrates, decision boundaries may incorrectly separate classes, especially for instances near class boundaries or in low-density regions far from learned distributions, producing stable yet incorrect classifications.
The \textit{scaling principle}~\cite{kaplan2020scaling} suggests that larger datasets and expressive architectures like deep neural networks should approximate true decision boundaries more closely. However, scaling cannot fully capture rare events in distribution long tails. Moreover, residual uncertainty can persists~\cite{hullermeier2021aleatoric}: \textit{aleatoric} uncertainty from intrinsic class ambiguity, and \textit{epistemic} uncertainty when instances fall outside well-represented training regions. This issue is exacerbated by incorrect labelling (Sec.~\ref{alg}) and inadequate feature representation (Sec.~\ref{dimred}). When labels are inconsistent or features fail to capture relevant dimensions, misclassification reflects not only model capacity but representational fidelity. Thus, even under apparent alignment, bias endures due to labelling imperfections, representational limits, or inherent uncertainty, revealing fundamental constraints of data-driven inference.\\

\begin{figure}[htbp]
    \centering
    \includegraphics[width=0.8\textwidth]{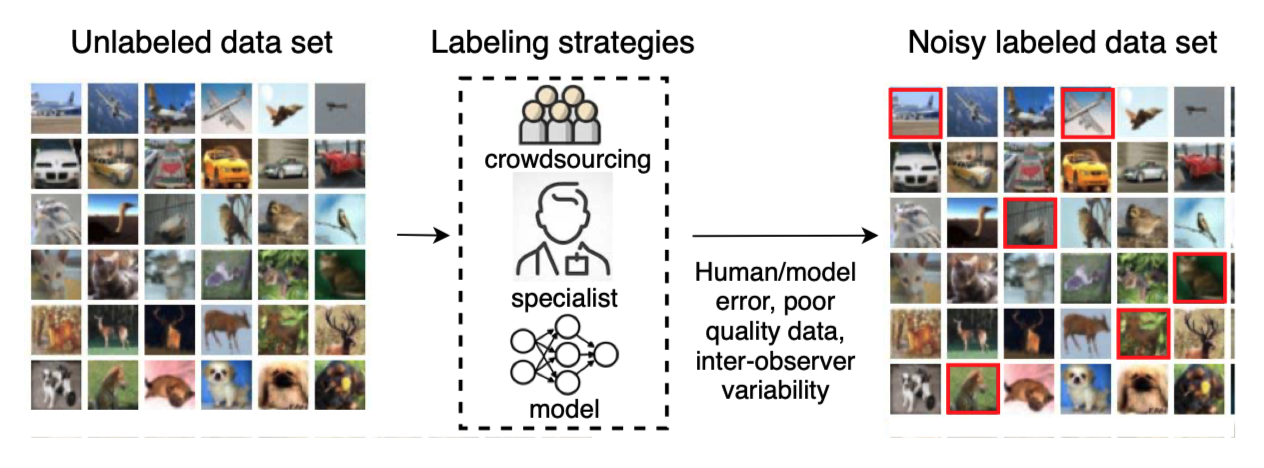} 
    \caption{Taken from~\cite{cordeiro2020survey}. Illustration of labeling errors and their impact on the alignment between system, data, and model. Incorrect or noisy labels can distort the learning process in two distinct ways. Examples belonging to the system are not correctly identified in the data, resulting in partial misalignment $D \cap S \neg M$. Examples are correctly included within $M \cap D \cap S$ but are associated with the wrong class, leading the model to learn systematically incorrect decision boundaries. Both forms of error illustrate how mislabeling propagates bias even when data quantity and model capacity appear sufficient.} 
    \label{fig:uncertLab} 
\end{figure}

\concept{C}{Incorrect or noisy labelling.}{mod:incnoi}  
When there is an imperfect correspondence between data instances and their true system states, the learning process is affected by labelling errors. These errors distort the alignment between the system, the data and the model in two main ways~\cite{northcutt2021confident}.
Firstly, when examples that exist in the system are not correctly identified or labelled in the data, a misalignment occurs. In this case, either relevant instances are omitted from the training set, $S \neg D \neg M$ or they are incorrectly marked as irrelevant $D \cap S \neg M$. This prevents the model from learning the true underlying associations.
Secondly, labeling noise may occur even when examples are correctly included in both the system and the data, corresponding to the region 
$M \cap D \cap S$, but are associated with an incorrect class or target value. In this case, the data provide misleading supervision, meaning the model learns patterns that are either systematically or randomly incorrect. This results in unreliable or biased predictions. As illustrated in Figure \ref{fig:uncertLab}, 
labeling errors can arise from various causes, such as human annotation mistakes (Sec.~\ref{label}), ambiguous or overlapping class definitions, inadequate domain expertise or inconsistencies in data integration processes. Regardless of their source, such errors undermine the fidelity of the learning process, propagating distortions throughout the model’s internal representations and subsequent inferences. \\

\begin{figure}[htbp]
    \centering
    \includegraphics[width=0.8\textwidth]{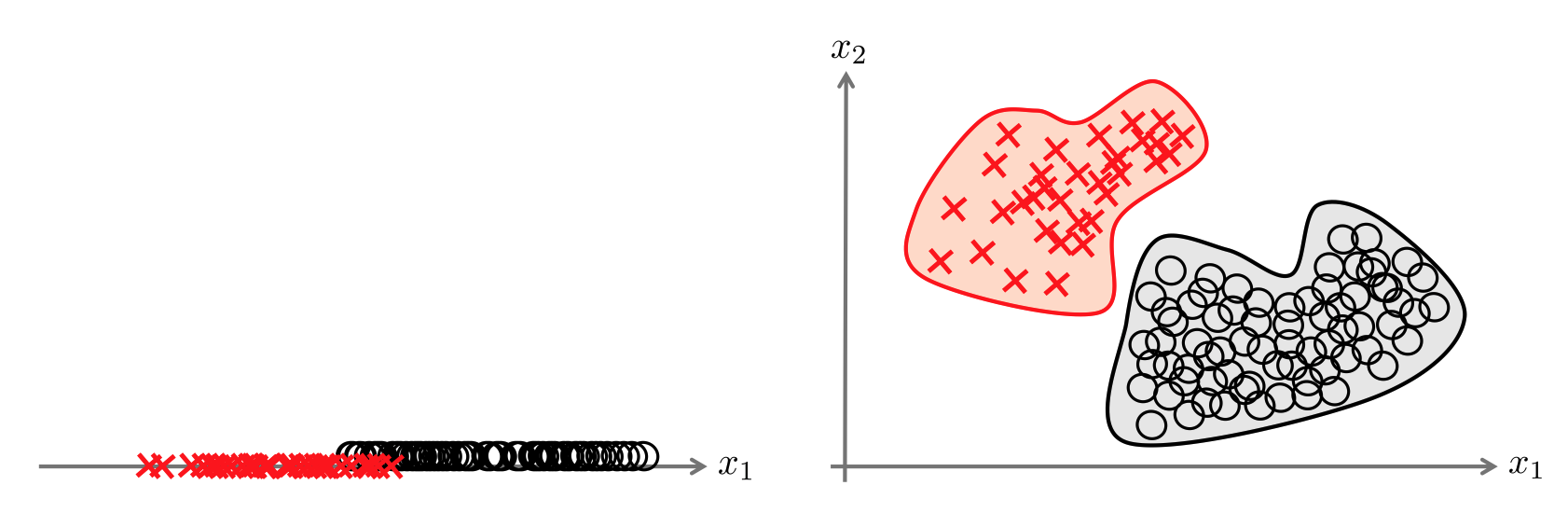} 
    \caption{Taken from~\cite{hullermeier2021aleatoric}. The figure shows how varying the number or scale of features affects class separability within the model’s decision space. When key features are missing or inadequately scaled, the resulting projection of the system into the data constrains the model, leading to distorted or overlapping decision boundaries.} 
    \label{fig:uncertFeat} 
\end{figure}

\concept{C}{Feature underrepresentation.}{mod:feaund}
When the features available in the data fail to capture the relevant dimensions of the system, the model learns from an impoverished or distorted representation of reality. As illustrated in Fig.~\ref{fig:uncertFeat}, varying the number or scale of features can substantially alter the separability of classes within the model’s decision space. If key explanatory variables are missing, aggregated or measured on the wrong scale, the decision boundaries inferred by the model may not align with the true structure of the system. This condition is primarily reflected in the region $D \cap S \neg M$. However, feature underrepresentation can also propagate distortions into other regions, for instance by inducing noise or hallucinations $M \cap D \neg S$ or systematic misclassification $M \cap D \cap S$.
Feature underrepresentation may be caused by measurement limitations, domain simplifications or design choices that restrict the range of the input space. It can also result from inadequate feature selection and scaling, which alters the relative influence of variables and reshapes the geometry of the decision surface (Sec.~\ref{dimred}).\\

\concept{C}{Data imbalance.}{mode:datimb}
When the data disproportionately represent certain classes, groups, or conditions of the system, the model learns a skewed approximation of the underlying distribution (Sec.~\ref{cimb}). In such cases, the learned decision boundaries are optimized toward the dominant classes, often at the expense of minority or infrequent ones. This situation can be described as a distortion primarily occurring within $ M \cap D \cap S$, where the sampling frequencies of system instances are unevenly reflected in the data, resulting in biased learning outcomes.
Data imbalance can manifest both at the class level—where one category vastly outnumbers others—and at the feature or subgroup level—where specific demographic, temporal, or contextual segments are underrepresented~\cite{10.1007/978-3-031-43415-0_20}.

The causes of data imbalance are diverse. They may arise from structural asymmetries in the system itself (e.g., historical biases or rare events), from sampling or data acquisition procedures that overlook minority categories, or from deliberate curation decisions that favor specific data sources~\cite{shah2025comprehensive}.\\

\concept{C}{Contextual or prompt bias.}{mod:conpro}
Unlike TAI, which maps predefined inputs to outputs within a fixed feature space, GenAI relies on contextual prompts to actively construct its outputs.  While this enables GenAI to generate responses that closely match the intended system, it also makes the model highly sensitive to the framing of input. When the prompt or context is misleading, incomplete or incorrectly specified, the model may produce outputs that differ from the true system — an effect analogous to feature underrepresentation, whereby critical information is absent or distorted.
Formally, this corresponds to the region $D \cap S \neg M$, in which the data and system contain relevant information, but the model fails to capture or interpret it due to contextual misalignment (Sec.~\ref{inter}). In GenAI, such misalignment often arises from prompt-induced bias stemming from ambiguous phrasing, narrow constraints or omitted context, which reduces the expressiveness of the input space and skews generation (Sec.~\ref{pind}).
While TAI is generally less context-dependent, it can also exhibit user-induced bias when its interface or intended use is misunderstood, leading to mis-specified inputs or misinterpretation of outputs. Therefore, in both paradigms, bias emerges not only from data and model limitations, but also from the dynamic interaction between human users, the model and the surrounding system context.
\\


\subsection{Validation and Evidence in Bias Verification}\label{sec:evidence}

From a design science perspective, bias verification repositions bias from a defect to a diagnostic tool in AI design. Operationalising this requires considering the evidential layers through which bias is identified, interpreted and managed.
TAI systems, with narrowly defined tasks and fixed inputs/outputs, enable straightforward internal validation using ground truth, stable distributions and standardised metrics (accuracy, precision, recall). However, internal validity deteriorate under changing conditions, as contextual shifts invalidate prior assurances.
GenAI systems complicate verification through open-ended outputs and probabilistic mechanisms. Their behaviour depends on prompts, context and evolving states, requiring broader, continuously updated evidence that encompasses correctness, adequacy, reliability and generalisability in dynamic environments.

In this regard, the distinction between internal and external validity offers a structured approach to linking the technical verification of AI systems with their contextual and ethical validation~\cite{mendling2025methodology}.\\

\concept{D}{Internal Validity}{val:internal} Concerns the extent to which a model’s observed behaviour can be causally attributed to its intended design — that is, its architecture, training data and optimisation procedures — rather than to uncontrolled or confounding factors. In the context of bias verification, internal validity establishes whether identified biases originate from identifiable mechanisms within the design process, such as data imbalance, representation error, or objective misalignment. It provides strong, albeit narrowly scoped, evidence of \textit{mechanistic integrity}, showing that the model behaves as designed and that its biases are traceable within the system’s internal logic.\\

\concept{D}{External Validity}{val:external}
By contrast, concerns the generalisability of the model’s behaviour and the persistence or transformation of bias across different contexts, datasets, and user interactions. It provides evidence of ecological reliability, showing whether the system remains robust, fair, and trustworthy once deployed. In this view, bias is not merely an internal design artefact, but a contextually mediated phenomenon that may be amplified, shifted, or attenuated when the system interacts with dynamic social or environmental conditions.

Taken together, these two forms of validity establish a hierarchy of evidence for verifying bias. Internal validity supports verification, confirming that the system’s mechanisms and internal biases are identifiable and reproducible, while external validity supports validation, ensuring that these mechanisms yield acceptable and reliable outcomes in situ. This dual framework aligns with the broader goal of design science: to mature AI systems through iterative cycles of controlled verification and contextual validation, thereby transforming bias into a constructive resource for reflective and accountable design.

\section{Illustrative Examples of AI Bias}\label{sec:exmp}

Although the theory of AI bias has been extensively studied, concrete examples are needed to reveal how it manifests and affects real-world outcomes. In this section, we present cases from various fields, including healthcare, recruitment, and generative content, to demonstrate how biased data, modelling choices, or representational errors can result in erroneous decisions, perpetuate stereotypes, and reinforce structural inequalities. These examples emphasise the importance of addressing bias throughout the entire AI lifecycle and demonstrate the practical consequences of unchecked algorithmic distortions in both TAI and GenAI systems.



\subsection{Social Bias}\label{SB}


Social bias is among the most insidious forms of distortion and originates from societal prejudices embedded in the data (e.g., stereotypes, systematic inequalities), as it reflects deep-seated prejudices in society, making it less apparent. This category of bias is not confined to a specific context but permeates human practices and the artifacts they produce. 
Unlike TAI systems that predict labels or numerical outputs, GenAI models generate text, images, or multimedia content, which makes biases more visible and impactful in narrative, visual, and contextual forms. In the following, some types of bias categorized as \textbf{Social Bias} are presented.
\subsubsection{Historical Bias} 
Historical bias refers to the embedding of past inequities or systemic discrimination into datasets and algorithms. It arises when data reflects historical injustices, thereby perpetuating these patterns in modern decision-making systems. 

\begin{biasbox}{Historical Bias}
\footnotesize{
\textbf{Source:} \conceptref{sou:socpre} \\[3pt]
\textbf{Modeling Stage:} \conceptref{life:datcol} and \conceptref{life:feaeng}\\[3pt]
\textbf{Modeling Factor:} \conceptref{mod:biadat}\\[3pt]
\textbf{Source of Evidence:} \conceptref{val:external} \\[3pt]

The model's behavior is described as residing in the region $M \cap D \cap \neg S$. The model learns the discriminatory patterns present in $D$, causing its outputs to deviate from the requirements of a fair system $S$.
\\

\textit{Example:} The model might learn that candidates with degrees from certain prestigious universities, a historically male-dominated field, or prior experience in certain male-dominated industries (such as tech startups) are more likely to be hired. If these characteristics are strongly correlated with male candidates, the algorithm may systematically favor male candidates even when gender is not explicitly used as a characteristic.
As a result, even without using gender as an explicit input, the model systematically favors male candidates, reproducing past inequities and reinforcing existing social biases.}
\end{biasbox}



\subsubsection{Survivorship Bias} 

It occurs when analysis focuses only on subjects that have ``survived'' a selection process, overlooking those that did not. This can lead to skewed conclusions because the missing data from non-survivors often contains critical insights~\cite{elston2021survivorship}. 

\begin{biasbox}{Survivorship Bias}
\footnotesize{
\textbf{Source:} \conceptref{sou:socpre} \\[3pt]
\textbf{Modeling Stage:} \conceptref{life:datcol} and \conceptref{life:feaeng} \\[3pt]
\textbf{Modeling Factor:} \conceptref{mod:misdat} \\[3pt]
\textbf{Source of Evidence:} \conceptref{val:external} \\[3pt]

The model's fundamental flaw stems from the $S \neg D \neg M$ region, where the critical information about failures is missing from both the data and the model's understanding. The model's output is based on a partial view of reality.\\

\textit{Example:}
Imagine that a study is conducted to identify the factors that contribute to a successful software startup, based solely on a dataset of companies that are still in business today that have managed to survive and thrive. However, this dataset ignores the many software companies that failed or were acquired and shut down.
Analysis would likely show that successful companies tend to have certain characteristics, such as a founding team with high technical expertise, significant venture capital investment, or a strong network within the industry. However, by looking only at surviving companies, the analysis ignores the many startups that had similar characteristics but ultimately failed. These failed companies may have faced unique challenges such as market timing, misalignment of product and customer needs, or poor management that didn't become apparent until after initial success. }

\end{biasbox}

\subsubsection{Subjectivity Bias} Occurs when personal opinions, beliefs, or preferences of individuals influence the data collection, interpretation, or decision-making process, leading to non-objective outcomes~\cite{doi:10.1126/science.185.4157.1124}. For instance, researchers may unintentionally interpret ambiguous results in ways that align with their hypotheses or expectations.

\begin{biasbox}{Subjectivity Bias}
\footnotesize{
\textbf{Source:} \conceptref{sou:socpre} and \conceptref{sou:humsub} \\[3pt]
\textbf{Modeling Stage:} \conceptref{life:datcol} and \conceptref{life:feaeng} \\[3pt]
\textbf{Modeling Factor:} \conceptref{mod:biadat} and \conceptref{mod:misdat} \\[3pt]
\textbf{Source of Evidence:} \conceptref{val:internal} and \conceptref{val:external} \\[3pt]

$S$ denotes the objective labels, while $D$ contains labels distorted by human subjectivity, introducing noise. Therefore Model learns this biased labeling pattern ($M \cap D \cap \neg S$).
In other situations, true labels may be absent from $S$ ($S \cap \neg D$) due to subjective interpretation. Consequently, $M$ cannot learn correct rules ($S \cap \neg D \cap \neg M$). \\

\textit{Example:} 
Consider
a company that is developing a sentiment analysis model to detect whether customer reviews are positive, neutral, or negative. They hire several human annotators to label a dataset of reviews. Regarding this review example, ``The product arrived late, but the customer service was very helpful”, Annotator A, who values punctuality, labels this as Negative; Annotator B, who prioritizes helpful service, labels it as Positive and; Annotator C chooses Neutral because the statement contains both positive and negative aspects.

The subjectivity of each annotator affects the label, and the model trained on this inconsistent labeling will inherit this subjectivity producing biased predictions in deployment.}

\end{biasbox}

\subsubsection{Cultural Bias}\label{calt} Arises when cultural norms, values, or perspectives influence the design, implementation, or interpretation of research, tools, or systems, leading to unfair advantages or disadvantages for specific cultural groups. 
For instance, Standardized Testing or contexts unfamiliar to certain cultural groups can disadvantage students not aligned with the dominant culture.

\begin{biasbox}{Cultural Bias}
\footnotesize{
\textbf{Source:} \conceptref{sou:socpre}, \conceptref{sou:humsub}\\[3pt]
\textbf{Modeling Stage:} \conceptref{life:datcol} and \conceptref{life:feaeng} \\[3pt]
\textbf{Modeling Factor:} \conceptref{mod:incnoi} \\[3pt]
\textbf{Source of Evidence:} \conceptref{val:external} \\[3pt]

The model's failure is rooted in $M \cap D \cap \neg S$. It has perfectly learned a culturally specific labeling schema that is misaligned with the full, cross-cultural reality of $S$.\\

\textit{Example:} In Medical Diagnosis we can point out the widely used depression diagnostic questionnaire that are developed and validated in Western cultures, where symptoms of depression often emphasize low mood and verbal expression of sadness. In some East Asian or African cultures, people with depression may report more physical symptoms (fatigue, body pain) rather than emotional symptoms. As a result, the tool may under-diagnose or misdiagnose depression in these populations, leading to inequitable access to care. Therefore, the diagnostic tool reflects the cultural assumptions of its creators and fails to generalize across diverse populations.}

\end{biasbox}

\subsubsection{Confirmation Bias} Confirmation Bias is the tendency to seek, interpret, and remember information that confirms one’s pre-existing beliefs or hypotheses while disregarding or undervaluing information that contradicts them. 
In TAI, confirmation bias occurs when developers, data scientists, or systems favor \textbf{Training data}, features, or outcomes that align with pre-existing assumptions or hypotheses, potentially leading to overfitting, poor generalization, or reinforcement of incorrect patterns~\cite{10.1007/978-3-031-43415-0_20, klayman1987confirmation}.

%
In GenAI instead, during \textbf{fine-tuning}, if annotators systematically favor responses that align with dominant cultural beliefs or commonly held opinions, the model learns to mimic these tendencies. This amplifies bias by rewarding conformity to majority perspectives rather than critical or balanced reasoning.

\begin{biasbox}{Confirmation Bias}
\footnotesize{
\textbf{Source:}  \conceptref{sou:humsub} \\[3pt]
\textbf{Modeling Stage:} \conceptref{life:datcol}, and \conceptref{life:feaeng} \\[3pt]
\textbf{Modeling Factor:} \conceptref{mod:misdat}, and \conceptref{mod:feaund}, and \conceptref{mod:conpro}\\[3pt]
\textbf{Source of Evidence:} \conceptref{val:internal} \\[3pt]

$S$ contains the full scope of information, including counterarguments, but the contradictory evidence is intentionally or unintentionally excluded from the Data ($S \cap \neg D$) due to the developer's or annotator's pre-existing beliefs. Therefore, $M$ cannot learn about these phenomena ($S \cap \neg D \cap \neg M$). 
In other cases, $S$ is defined by a complete set of predictive features but $D$ and subsequent model training are restricted to a narrow, subjectively chosen subset of features ($D \cap S \cap \neg M$).\\

\textit{Example 1:} A data scientist is building a fraud detection model based on the belief that high transaction amounts are the most important indicator of fraud. They prioritize features related to transaction size and ignore other potential signals such as geographic patterns or unusual login behavior.
During evaluation, they select validation examples that showcase the model catching high-amount frauds, ignoring cases where it fails on smaller but sophisticated frauds.\\ 

\textit{Example 2:} A conversational AI chatbot is deployed on a news platform to engage users in political discussions. A user prompts the chatbot with:

\texttt{
``Why are environmental regulations harmful to the economy?"}

The model responds with a detailed explanation supporting the assumption that environmental regulations hurt economic growth, citing examples like compliance costs for businesses. It does not offer a counterargument, such as the long-term economic benefits of sustainable practices or green innovation.
}
\end{biasbox}

\subsubsection{Anchoring Bias} Anchoring bias is a cognitive bias where an individual relies too heavily on an initial piece of information when making decisions or judgments, even when subsequent information suggests the anchor is irrelevant or incorrect. It can
affect model performance due to over-reliance on default or initial parameters,
influence feature selection or weighting, especially if initial assumptions are incorrect, or
skew results in iterative labeling or reinforcement learning~\cite{rhue2023anchoring}.

\begin{biasbox}{Anchoring Bias}
\footnotesize{
\textbf{Source:}  \conceptref{sou:humsub} \\[3pt]
\textbf{Modeling Stage:} \conceptref{life:datcol}, 
\conceptref{life:feaeng}
 \\[3pt]
\textbf{Modeling Factor:} \conceptref{mod:biadat}, 
\conceptref{mod:misdat}
\\[3pt]
\textbf{Source of Evidence:} \conceptref{val:external} \\[3pt]

The initial, anchored assumptions (e.g., ``age/income are most important'') cause the team to create a skewed dataset ($D$) by over-weighting certain features.

$M$ learns from this artificially biased data, resulting in a suboptimal and unfair model that is misaligned with $S$ ($S \cap \neg D \cap \neg M$).\\

\textit{Example:} Consider a team is building a machine learning credit scoring model and initially assumes that age and income are the most important predictors of creditworthiness, based on previous industry practices. They select and overweight features related to these two variables early in development. Even when exploratory data analysis later suggests that debt-to-income ratio or recent payment history are stronger predictors, the team sticks to their initial assumptions because their feature selection and model tuning were anchored to these early beliefs.}

\end{biasbox}

\subsubsection{Group Attribution Bias} Occurs when individuals attribute the behavior or characteristics of a single member of a group to the entire group or vice versa. It reflects a tendency to generalize based on limited information and often leads to stereotyping. Assuming all members of a demographic group share a particular trait because one prominent individual does is a common example of this type of bias. In AI, models trained on biased data may unfairly attribute behaviors or characteristics to groups, leading to discriminatory outcomes (e.g., predictive policing models disproportionately targeting minority communities).

\begin{biasbox}{Group Attribution Bias}
\footnotesize{
\textbf{Source:}  \conceptref{sou:humsub} \\[3pt]
\textbf{Modeling Stage:} \conceptref{life:datcol}, 
\conceptref{life:feaeng}
 \\[3pt]
\textbf{Modeling Factor:} \conceptref{mod:biadat} \\[3pt]
\textbf{Source of Evidence:} \conceptref{val:external} \\[3pt]

$D$ is a historical record skewed by societal prejudices and over-policing, creating a false correlation between features such as demographics and crime rate.
$M$ then learns this spurious correlation from the data ($M \cap D \cap \neg S$). 

\textit{Example:}\\
A predictive policing AI system is trained on historical crime data that disproportionately reflects over-policing in certain neighborhoods, often with higher populations of minority communities. Therefore, the model learns to associate crime risk with demographic or geographic features. As a result, the system continually flags these communities for higher surveillance, even when actual crime rates do not justify it.

This leads to a self-reinforcing loop where more police presence leads to more reported incidents, which then feeds back into the model. This model unfairly attributes risk to an entire group or area based on biased historical data.
}
\end{biasbox}

\subsubsection{Echo Chamber Bias}

The Echo Chamber effect in GenAI occurs when a model repeatedly reinforces or amplifies a user’s opinions, tone, or reasoning during multi-turn interactions. This happens because conversational AI systems are optimized to maintain coherence and user satisfaction, often by aligning with the user’s assumptions instead of challenging them or introducing alternative perspectives. 
Unlike one-shot confirmation bias, which occurs in a single response, the Echo Chamber effect compounds over multiple conversational turns, gradually strengthening the user’s initial stance and creating a sense of validation.


\begin{biasbox}{Echo chamber Bias}
\footnotesize{
\textbf{Source:}  \conceptref{sou:humsub} \\[3pt]
\textbf{Modeling Stage:} \conceptref{life:datcol}, and 
\conceptref{life:feaeng}
\\[3pt]
\textbf{Modeling Factor:} \conceptref{mod:biadat} \\[3pt]
\textbf{Source of Evidence:} \conceptref{val:external}\\[3pt]

$S$ is the balanced, nuanced, and evidence-based view on a topic.
$D$ in this context is the conversation history, which becomes increasingly skewed and one-sided due to the user's repeated biased inputs.
$M$ is trained to be context-aware and aligns its responses with this progressively more biased conversational data.
Therefore, the model's behavior is a direct result of learning from and adapting to the biased data of the conversation thread, placing it in the $M \cap D \cap \neg S$ region.

\textit{Example:}  A user interacts with a chatbot about immigration policy:

\texttt{User: ``Immigrants are taking away jobs from locals, right?”}

\texttt{Model: ``Many people believe immigration can impact job availability. Some argue that competition increases.”}

\texttt{User: ``Exactly! And that’s why immigration should be reduced.”}

\texttt{Model: ``Yes, reducing immigration is often suggested as a way to protect local employment.”}

Over time, the model mirrors and strengthens the user’s perspective, creating an echo chamber effect, even though economic research provides more nuanced views.
}
\end{biasbox}


\subsubsection{Framing Bias}
In GenAI, Framing Bias refers to the bias introduced by how a task, prompt, or dataset is structured or contextualized, which influences what the model learns to generate and how it generates it. It reflects the perspective or assumptions embedded in the framing, whether by developers, annotators, or users which can steer the model toward particular narratives, interpretations, or tones. Framing bias in GenAI occurs when the wording, structure, or presentation of input data and prompts leads the model to favor certain perspectives, values, or interpretations while ignoring or downplaying others. 
Framing bias can happen during \textbf{Prompt Framing} by Users
, \textbf{Annotation and Feedback Framing}, \textbf{System Prompt Framing (Instruction Tuning)}.

\begin{biasbox}{Framing Bias}
\footnotesize{
\textbf{Source:}  \conceptref{sou:humsub} \\[3pt]
\textbf{Modeling Stage:} \conceptref{life:datcol}, 
\conceptref{life:feaeng}
\\[3pt]
\textbf{Modeling Factor:} \conceptref{mod:conpro} \\[3pt]
\textbf{Source of Evidence:} \conceptref{val:external} \\[3pt]

$S$ is the complete, multi-faceted reality of the topic (e.g., both positive and negative impacts of social media). $D$ is the user's prompt, which presents a framed, narrow, and subjective slice of the system. $M$ is highly sensitive to this input context and generates an output that aligns with the framed perspective model. Therefore, the core failure is that the input data provides a biased context that misrepresents the full system. This leads the model to produce an output that, while aligned with the prompt, is misaligned with $S$ ($M \cap D \neg S$).\\

\textit{Example:} Compare the following two prompts: 

\texttt{Prompt A: ``Explain why social media is bad for teenagers.''}\\
\noindent
 The model focuses more on negative effects (addiction, anxiety).

\texttt{Prompt B: ``Discuss the impact of social media on teenagers.''}\\
\noindent
The model presents both pros (connection, self-expression) and cons (Body Image Issues, Privacy and Safety Risks).
}

\end{biasbox}



\subsection{Data Representation Bias}\label{RB}
 
Representation bias can arise once the data is 
imbalanced or the representation of populations or scenarios is incomplete. This skewed sampling leads directly to Data Coverage Bias, where the final insights or prediction models are not generalizable to the full population. The trained models on these dataset are biased, underperforming, or unfair when applied to underrepresented groups or scenarios. An illustrative example is found in urban traffic datasets, which frequently overlook the needs of individuals with disabilities. This oversight leads to AI-driven solutions that fail to consider their experiences and requirements, rendering systems less inclusive.

Data Representation Bias in GenAI refers to systematic distortion caused by how information is encoded, structured, and prioritized within the training data. This category of bias influences what the model learns as ``important'' and how concepts are represented internally. 

Representation bias occurs before and during training, due to several reasons such as: \textbf{Imbalanced token frequencies} such as some words, cultural concepts, or dialects occur more often, \textbf{Data formatting inconsistencies} such as formal vs. informal language dominates and, \textbf{Hierarchical structures} in text like headlines prioritized over body text in scraped datasets.


\subsubsection{Data Coverage (Selection) Bias}\label{cov} Data Coverage Bias occurs when the training data lacks sufficient diversity or completeness across the full range of concepts, languages, contexts, or scenarios that the model is expected to handle. Unlike selection bias (which is about how data is sampled), coverage bias focuses on gaps or omissions in the dataset that lead to systematic blind spots in the model’s knowledge or generation capabilities.
This type of bias may occur during \textbf{Data Collection}, once the corpus emphasizes certain domains but underrepresents others; at the \textbf{Preprocessing Stage}, where rare or low frequency data may be pruned during deduplication or cleaning steps, removing underrepresented voices while retaining dominant narratives; or eventually at the \textbf{Model Training Stage} even if included, underrepresented data may not influence the model effectively due to low frequency compared to dominant categories.

In the context of GenAI, this bias leads models to produce outputs that reflect dominant patterns in the training corpus while underrepresenting or misrepresenting minority voices, cultural contexts, and low-resource languages. For example, a language model trained primarily on English-language Western media may produce fluent and contextually rich responses for Western topics but struggle with non-Western cultural references, low-resource languages, or marginalized community perspectives. This not only reduces the inclusivity and generalizability of the model but also risks reinforcing existing societal inequalities by privileging mainstream content and neglecting global diversity. 
Data coverage bias contributes to \textit{hallucination} once there is a Lack of Grounded Information. If a model has never or rarely seen data about a specific topic or demographic, it guesses or fabricates responses using patterns from unrelated but more common data or produces outputs that are plausibly sound but factually incorrect or nonsensical. It may also occur by \textit{overgeneralization} from dominant data when certain domains dominate the training data, and the model overapplies learned associations to unfamiliar contexts. In addition, inadequate representation of edge cases may also lead the model to generate typical or averaged responses, and hallucinated content.
Once data coverage is thin, the model may use \textit{statistical interpolation} to fill in gaps (Uncertainty Filling), leading to ``confidently wrong" outputs. 

    


\begin{biasbox}{Data Coverage Bias}
\footnotesize{
\textbf{Source:}  \conceptref{sou:humsub} \\[3pt]
\textbf{Modeling Stage:} \conceptref{life:datcol}, 
\conceptref{life:feaeng}
 \\[3pt]
\textbf{Modeling Factor:} \conceptref{mod:misdat} and \conceptref{mode:datimb}\\[3pt]
\textbf{Source of Evidence:} \conceptref{val:internal} \\[3pt]

$S$ includes the entire population and all relevant scenarios (e.g., both urban and rural patients, all language dialects). $D$ provides only a partial, non-representative sample of $S$, systematically omitting entire subpopulations or contexts and $M$ is trained on this incomplete view. Therefore, the core failure is the $S \cap \neg D \cap \neg M$ region.\\

\textit{Example:} In medical research, using data primarily from urban hospitals can introduce sampling bias, as the results may not generalize to rural populations
who may have different lifestyle factors, environmental exposures, and healthcare access. As a result, the AI model performs poorly on rural patients, leading to inaccurate risk assessments and potential health disparities. \\

}

\end{biasbox}


\subsubsection{Bias by Class Imbalance}\label{cimb} Bias by Class Imbalance arises when certain classes or categories are significantly underrepresented in the training data, often due to historical inequities, systemic discrimination, or unequal access to the systems or environments that generate data. This imbalance can lead to models that are biased toward the majority class, systematically misclassifying or overlooking minority classes.
It reflects not just technical flaws, but often deep-rooted social inequalities embedded in the data.

\begin{biasbox}{Class Imbalance Bias}
\footnotesize{
\textbf{Source:}  \conceptref{sou:humsub}, 
\conceptref{sou:teclim}
 \\[3pt]
\textbf{Modeling Stage:} \conceptref{life:datcol}, and 
\conceptref{life:feaeng}
 \\[3pt]
\textbf{Modeling Factor:} \conceptref{mode:sysmis}
 \\[3pt]
\textbf{Source of Evidence:} \conceptref{val:internal} \\[3pt]

$S$  contains multiple classes in their natural distribution. $D$ provides complete sample but where one or more classes are significantly underrepresented. 
$M$ is trained on this skewed data and optimized for the majority class(es), and its performance and decision boundaries are systematically biased against the minority class(es). This creates a gap not in the way classes are learned by the system  ($S \cap D \cap M$).\\

\textit{Example:}  
    In skin cancer detection models, training datasets often contain more images of lighter skin tones because historically, medical imaging and research have focused on white patients. The model performs well on lighter skin tones but is less accurate on darker skin, leading to disparities in early detection and treatment. In this example, the class imbalance reflects historical underrepresentation and leads to unfair outcomes for marginalized groups.
}
\end{biasbox}


\subsubsection{Dimensionality Reduction (Data Transformation)}\label{dimred}
Data Representation Bias may happen in the Data Transformation or Data Encoding for the downstream algorithms' usage. 
\textbf{Dimensionality Reduction} step may cause loss of critical information relevant to specific groups while emphasizing less important dimensions. 
Representation bias can emerge in the \textbf{feature extraction} step, in selecting the features that favor certain groups or neglect important attributes for others. Further more, in the feature selection stage, by Selection of features that encode implicit biases.

\begin{biasbox}{Data Transformation Bias}
\footnotesize{
\textbf{Source:}  
\conceptref{sou:teclim}
 \\[3pt]
\textbf{Modeling Stage:} \conceptref{life:datcol},
\conceptref{life:feaeng}
 \\[3pt]
\textbf{Modeling Factor:} \conceptref{mod:feaund} \\[3pt]
\textbf{Source of Evidence:} \conceptref{val:internal} \\[3pt]

$S$ and $D$ contain features critical for certain groups. The dimensionality reduction algorithm acts as a flawed filter, discarding or down-weighting these features during the creation of the model's input representation. Thus, $M$ never receives this critical information ($D \cap S \cap \neg M$).\\

\textit{Example:} Once  PCA or similar methods introduce artifacts based on prior distribution assumptions, they may downweight features that are crucial for underrepresented or minority groups, leading to biased representations and degraded model performance for those populations.
In medical diagnosis, rare but important symptoms more common in minority populations may be excluded or diminished in reduced feature spaces, leading to misdiagnosis or unequal treatment recommendations.
}

\end{biasbox}

\subsubsection{Data Encoding Bias}

Representation bias may also occur when \textbf{embedding spaces} of textual contents or images preserve or amplify social, cultural, or demographic stereotypes present in the training data. This transformation may reflect a particular tendency while neglecting others, causing bias against underrepresented instances in the latent space. For example, techniques that represent words or phrases through numerical values can emphasize common linguistic associations while ignoring less frequent contexts or meanings~\cite{mohammadi2025identifying}. Similarly, relying on aggregate statistics, such as averages, can obscure differences between majority and minority groups, reducing the system's ability to capture diversity.

\begin{biasbox}{Data Encoding Bias}
\footnotesize{
\textbf{Source:}  \conceptref{sou:socpre}, \conceptref{sou:teclim} \\[3pt]
\textbf{Modeling Stage:} \conceptref{life:feaeng}, 
\conceptref{life:pretra}
\\[3pt]
\textbf{Modeling Factor:} \conceptref{mod:feaund} \\[3pt]
\textbf{Source of Evidence:} \conceptref{val:internal} \\[3pt]

$M$ learns vector representations (embeddings) from $D$ that capture and amplify societal stereotypes (e.g., gender occupations). These learned embeddings are a core part of the model's   ``understanding'' but encode a distorted view that is misaligned with a fair system ($M \cap D \cap \neg S$). This is then reflected in the downstream generated outputs.\\
\textit{Example:}     In word embeddings (e.g., Word2Vec, GloVe, BERT), numerical representations of words can preserve or even amplify social and cultural stereotypes. For instance, embeddings may associate ``nurse'' more closely with   ``female'' and ``doctor'' with ``male'' based on biased language patterns in the training data. As a result, downstream applications like chatbots, translators, or resume screeners may perpetuate these stereotypes. 
}

\end{biasbox}

GenAI models often inherit biases from their underlying embedding representations. 
Embedding Bias can emerge in different contexts of GenAI models:

\begin{itemize}
    \item Static Word Embeddings Bias: Word2Vec and GloVe capture co-occurrence statistics from large corpora, often reflecting biased social norms. This context is in common with TAI and NLP, earlier discussed extensively.
    \item Contextual Embeddings Bias: Transformer-based models such as BERT and GPTs, use dynamic embeddings that change based on context, making bias detection and mitigation more complex. For example,  A GPT model is asked to complete: ``The programmer finished the project. (...) went to the conference...''. 
The model outputs ``He'', relying on a statistically dominant ``programmer $\rightarrow$ male'' association in its contextual embeddings, despite no gender specification in the prompt.  
    \item Multimodal Embeddings Bias: Models like CLIP (Contrastive Language–Image Pretraining) align text and image embeddings but may reinforce visual stereotypes or cultural imbalances. For example, a fashion retailer's multimodal model, when prompted with ``business casual outfit'', predominantly generates images of light-skinned models in Western attire, as its embeddings were trained on internet data that underrepresents non-Western fashion and darker skin tones.    
    \item Overcompression Bias: During embedding formation, models compress high-dimensional inputs into low-dimensional spaces. Frequent patterns dominate the representational space, while rare or minority features are ``averaged out” or inaccurately clustered. 
    For example,  in a language model's embedding space, dialects, minority identities, or cultural concepts with low token frequency are clustered inaccurately or lose their distinct meaning, leading to higher rates of misrepresentation or hallucination when generating content related to these topics.

\end{itemize}

\subsection{Measurement and Decision Bias}\label{MDB}

Measurement Bias or Instrument Bias can  arises when the measurement tools or instruments used in research systematically favor certain outcomes or groups, leading to inaccurate data or skewed results. On the other hand, Decision bias refers to the logical decisions made by a data collector, a developer or systematically by model in the training process.

GenAI systems also inherit or amplify measurement and decision biases that arise at different stages of \textbf{Data Collection}, \textbf{Model Training}, and \textbf{Deployment}. The intersection of measurement bias and decision bias in GenAI systems presents critical challenges for fairness, accountability, and reliability. 
While measurement bias provides distorted inputs, such as biased representations of gender roles, Decision bias amplifies or filters outputs based on subjective preferences during fine-tuning or inference.



\noindent
\subsubsection{Measurement Bias}\label{meas}

This can occur due to poorly calibrated instruments, differing interpretations, or inappropriate tools for the population under study. 
In the GenAI context, by \textbf{Instrument Bias} we refer mainly to those biases imposed by systemic distortions introduced by \textbf{technical artifacts}, including software libraries, tokenizers, filters, annotation tools, quality metrics, and other engineering decisions. For example, at \textbf{Data Collection} stage, a web scraper collects mostly text from Western news outlets and forums, underrepresenting voices from the Global South or non-English communities. 
\textbf{Tokenization tools} used are often trained on dominant languages or corpora. They split minority-language words or dialects into multiple sub-tokens, increasing length and reducing semantic clarity.
\textbf{Automated quality scoring tools}, such as perplexity, readability, and classifier-based filters, often favor mainstream grammar and vocabulary and lead to a biased ranking of data. For example, a data deduplication pipeline keeps formal essays but removes spoken word transcripts or nonstandard poetry. 

\begin{biasbox}{Measurement Bias}
\footnotesize{
\textbf{Source:}  \conceptref{sou:teclim} \\[3pt]
\textbf{Modeling Stage:} \conceptref{life:datcol}, \conceptref{life:feaeng} \\[3pt]
\textbf{Modeling Factor:} \conceptref{mod:biadat}  \\[3pt]
\textbf{Source of Evidence:} \conceptref{val:internal}
 \\[3pt]

The flawed measurement process creates a dataset $D$ that is a systematically distorted representation of $S$. $M$ is then constrained to learn from this flawed $D$. The resulting model's behavior is inherently misaligned with the true system ($M \cap D \cap \neg S$).\\

\textit{Example 1:} Using blood pressure cuffs designed for adults on children can produce inaccurate data generated by inappropriate instruments, in the  upcoming studies. \\

\textit{Example 2:} Consider a LLM like GPT or LLaMA trained with a tokenizer designed primarily for English and other high-resource languages.
When the model encounters a Swahili sentence, the tokenizer often splits words into multiple sub-tokens because it was not optimized for Swahili morphology. For example:
``Ninapenda kusoma vitabu" (``I like reading books")
might be split into [Ni, nap, enda, ku, soma, vi, tabu], while an equivalent English sentence is tokenized as [I, like, reading, books]. The longer sub-token sequence increases the computational ``cost'' of Non-English text, discouraging its representation during training and generation. As a result, the model may produce less coherent or lower-quality Swahili outputs.
}

\end{biasbox}

\subsubsection{Labeling Bias}\label{label} 
Labeling Bias occurs when data annotators introduce biases, consciously or unconsciously, during the labeling process or feedback mechanism, which subsequently influence how the model interprets, learns, and generates content. 


\textbf{Labeling tools}, \textbf{interfaces}, and \textbf{annotation instructions} play a crucial role in shaping annotator decisions, often reinforcing specific social norms. For example, under vague or poorly defined labeling guidelines, annotators might flag terms such as ``queer pride” as political or inappropriate, even when these terms are not inherently offensive.

This issue is particularly significant in GenAI because labeled data is central to \textbf{fine-tuning}, \textbf{supervised learning}.

\begin{biasbox}{Labeling Bias}
\footnotesize{
\textbf{Source:}  \conceptref{sou:humsub}, \conceptref{sou:teclim} \\[3pt]
\textbf{Modeling Stage:} \conceptref{life:datcol}, and 
\conceptref{life:feaeng}
 \\[3pt]
\textbf{Modeling Factor:} \conceptref{mod:incnoi}, and \conceptref{mode:sysmis} \\[3pt]
\textbf{Source of Evidence:}  
\conceptref{val:external}
 \\[3pt]

The labels in $D$ are systematically incorrect due to human or technical error, providing flawed supervision.
$M$ learns these incorrect patterns from $D$, causing its internal representations and outputs to be biased and misaligned with $S$ ($M \cap D \cap \neg S$) or aligned but following a wrong classification ($M \cap D \cap S$).

\textit{Example:} Labeling images of engineers disproportionately as male due to societal stereotypes may introduce labeling bias to the training dataset. The resulting model learns to associate ``engineer” with ``male”, perpetuating the stereotype and potentially misclassifying images of female engineers. This flaw in labeling is driven by implicit social biases of annotators.
}
\end{biasbox}

\subsubsection{Algorithmic Bias}\label{alg}

Algorithmic bias is linked to errors that \textbf{decision logic} applied by an algorithm, which can introduce significant distortions if \textbf{evaluation criteria} are inadequately chosen or implicitly reflect biases present in training data, flawed design or biased decision-making process. For example, automated grading systems often use clustering algorithms to group students with similar characteristics. If these algorithms fail to consider socioeconomic and cultural factors that influence student performance, they risk mislabeling individuals as less capable, perpetuating cycles of inequality.

Some forms of Algorithmic Bias are as follow:

\begin{itemize}
    \item \textbf{Optimization Bias} occurs during \textbf{training procedure} once a model's objective function, constraints, or evaluation metrics are improperly chosen or overly simplified, leading to suboptimal or biased results that do not align with real-world fairness or utility goals. Models optimized for ``coherence'' or ``realism'' may prioritize majority viewpoints (e.g., generating ``nurse'' as female). Poor Metric Selection and Local Optima could be among the possible causes of Optimization Bias.

\begin{biasbox}{Optimization Bias}
\footnotesize{
\textbf{Source:}  \conceptref{sou:teclim} \\[3pt]
\textbf{Modeling Stage:}\conceptref{life:modtra} \\[3pt]
\textbf{Modeling Factor:} \conceptref{mode:sysmis} \\[3pt]
\textbf{Source of Evidence:} \conceptref{val:internal} \\[3pt]

$M$ s successfully learning the patterns in $D$, but the chosen objective function (e.g., overall accuracy) causes it to optimize for a goal that is misaligned with the real-world system's requirements for fairness or utility $(M \cap D \cap \neg S)$.\\

\textit{Example:} In a hiring system, a model that is optimized for metrics like accuracy, precision, or recall while excluding fairness constraints, such as equal opportunity, from the optimization process in a hiring algorithm optimizing for efficiency may unintentionally prioritize applicants from a privileged group.  
}

\end{biasbox}

    \item \textbf{Thresholding Bias} in AI models typically emerges during the decision-making stage, particularly when a threshold is applied to the output probabilities or scores of a model to make a final classification or prediction. This bias can be introduced, amplified, or mitigated depending on how the threshold is chosen and how it interacts with imbalanced data or population subgroups. Thresholding Bias can emerge at \textbf{Post-Training/Inference Stage}, where a model (e.g., a classifier) outputs a probability score and a threshold is used to decide class membership. At this stage, the model has been trained and is now generating predictions (scores or probabilities). 
    This typically happens during \textbf{evaluation}, or \textbf{validation} steps on hold-out datasets before deployment. Bias can emerge if thresholds are optimized for overall performance without fairness considerations. A \textit{global threshold} of $\mathbf{x}$ might yield $90\%$ precision overall, but only $70\%$ for a minority group. Also, different subgroups may have different score distributions after training. A single threshold may lead to different error rates (e.g., higher false negatives for one group). 
    Applying the same threshold without accounting for different costs of false positives/negatives across groups (Asymmetric cost of errors) can lead to thresholding bias. 

    Thresholding bias may also arise due to a lack of \textit{subgroup calibration}, where a model’s probability estimates are not calibrated equally well across groups, a global threshold leads to inconsistent decisions. 
    Also, Thresholding Bias could become evident at the \textbf{Deployment Stage}, where the model is now live and making decisions in real-world scenarios. 	
    
\begin{biasbox}{Thresholding Bias}
\footnotesize{
\textbf{Source:}  \conceptref{sou:humsub}, \conceptref{sou:teclim} \\[3pt]
\textbf{Modeling Stage:}\conceptref{life:modtra}, \conceptref{life:dep}, 
\conceptref{life:posdep}
 \\[3pt]
\textbf{Modeling Factor:} \conceptref{mod:biadat}, \conceptref{mode:sysmis} \\[3pt]
\textbf{Source of Evidence:} \conceptref{val:external} \\[3pt]

The core model exists in the aligned space ($M \cap D \cap S$), but the final decision rule (the threshold) applied to its outputs can create misalignment. A single threshold may work for the majority in the data but fail for subgroups, effectively pushing their outcomes into the $M \cap D \cap \neg S$ region.\\

\textit{Example:}  In a hiring model, women and men might receive different score distributions, leading to fewer women being selected if the threshold is not calibrated for fairness.\\

}

\end{biasbox}

    \item \textbf{Overfitting Bias} refers to the systematic error introduced during \textbf{Training/Optimization} when a model is too complex relative to the amount or diversity of the training data or once the choice of hyperparameter, architectures, and features are not validated correctly, tuning for accuracy without considering generalizability. 
    The overfitted model memorizes the training data rather than learning to generalize. The bias here is not in the sense of model bias-variance tradeoff, but in the sense of biased performance across groups due to overfitting. 
    \textbf{Data Leakage} may also lead to overfitting, where the	information from the test set leaks into the training process.

\begin{biasbox}{Overfitting Bias}
\footnotesize{
\textbf{Source:}  \conceptref{sou:teclim} \\[3pt]
\textbf{Modeling Stage:}\conceptref{life:modtra}
 \\[3pt]
\textbf{Modeling Factor:} \conceptref{mod:biadat} \\[3pt]
\textbf{Source of Evidence:} \conceptref{val:external} \\[3pt]

$M$ memorizes spurious correlations and noise specific to $D$ but fails to capture the underlying principles of the real-world system ($M \cap D \cap \neg S$), leading to poor generalization.\\

\textit{Example:} A tech company develops a deep learning model to predict candidate success (e.g., retention and performance) based on historical hiring data. The goal is to automate parts of the hiring process and shortlist promising applicants.
The training dataset consists largely of past hires, who are predominantly male and come from a limited number of universities and regions. The model is overparameterized (such as a deep neural network with many layers) and is trained to maximize accuracy. Without proper regularization or subgroup validation, the model starts to memorize patterns associated with the dominant group, like ``success" correlates with graduating from a certain university or having certain keywords common in male resumes.

}

\end{biasbox}

    \item \textbf{Inductive Bias} refers to the assumptions made by a learning algorithm to generalize beyond the training data. These biases guide the model in selecting hypotheses and influence how it interprets unseen data, shaping its ability to generalize effectively. Unlike overfitting or thresholding bias, which are runtime or post-training issues, inductive bias is built-in early, and it guides how the model will behave during training and prediction. For example, at the \textbf{Model training} stage, if one choose a 
    Convolutional Neural Networks (CNNs), it assumes a spatial locality inductive bias, making them well-suited for image data or the Decision Tree algorithms that implicitly assume the data can be split hierarchically, which is an inductive bias of the algorithm or choosing the gradient-based methods that assumes smoothness of the loss function. So that this type of bias is not necessarily harmful or a sign of error, but it is a built-in assumption of the model. 

\begin{biasbox}{Inductive Bias}
\footnotesize{
\textbf{Source:}  \conceptref{sou:teclim} \\[3pt]
\textbf{Modeling Stage:}\conceptref{life:modtra}
 \\[3pt]
\textbf{Modeling Factor:} \conceptref{mode:sysmis}\\[3pt]
\textbf{Source of Evidence:} \conceptref{val:internal} \\[3pt]

A misaligned inductive bias (e.g., assuming linearity in a nonlinear system) acts as a fundamental constraint. It prevents $M$ from ever learning certain true aspects of the system, even if they are present in the data. The required solution exists $S$, but the model's built-in assumptions block it from being learned ($S \cap D \cap \neg M$).\\

\textit{Example:}    Imagine a hospital develops a predictive model to identify patients at risk of hospital readmission within 30 days, using features such as age, diagnosis codes, length of stay, and past medical history. The data science team initially uses a linear logistic regression model due to its interpretability and simplicity. This model assumes additive and linear relationships among features for instance, it assumes that each additional year of age contributes a fixed increase to readmission risk. However in reality, patient risk is influenced by complex, nonlinear interactions such as comorbidity combinations or nonlinear thresholds in biomarker levels that the linear model cannot capture. As a result, the model underestimates risk for patients with certain chronic conditions that interact in a multiplicative, nonlinear manner. As a consequence, this model missed high-risk patients, particularly among groups with complex health profiles exacerbating the healthcare disparities. 
}

\end{biasbox}

    \item \textbf{Reward Signal Bias}  In GenAI, Reward Signal Bias refers to the bias introduced during the fine-tuning phase, especially in \textbf{Reinforcement Learning from Human Feedback (RLHF)}, where models are trained to optimize for human preferences. 
    Reward Signal Bias occurs when the feedback or scoring system used to fine-tune a model reflects skewed, subjective, or culturally dominant values. It may emerge due to Biased annotators' preferences, unclear or narrow reward criteria, overemphasis on fluency or persuasiveness and, homogeneity in feedback sources once raters mostly come from similar backgrounds. 

\begin{biasbox}{Reward Signal Bias}
\footnotesize{
\textbf{Source:}  \conceptref{sou:humsub}, \conceptref{sou:teclim} \\[3pt]
\textbf{Modeling Stage:}\conceptref{life:finepro}
 \\[3pt]
\textbf{Modeling Factor:} \conceptref{mod:biadat} \\[3pt]
\textbf{Source of Evidence:} \conceptref{val:external} \\[3pt]

$M$ may be capable of generating system-aligned outputs, but the reward signal during fine-tuning reinforces patterns that are prevalent in the human feedback data but are misaligned with the true, broader requirements of the system ($ M \cap D \cap \neg S$), such as fairness and balance.\\

\textit{Example:} During RLHF for a conversational AI, annotators consistently rank sycophantic responses higher than factually correct but challenging ones. For the prompt "Is communism a viable economic system?", outputs that vaguely agree with the user's pre-existing political stance receive high rewards, while balanced analyses of pros/cons receive low rewards. 
}

\end{biasbox}
    \item \textbf{Benchmarking and Evaluation Bias} refers to the systematic distortion in model assessment that arises from the use of flawed, narrow, or non-representative benchmarks, evaluation metrics, or testing procedures. 
    This bias does not exist in the model itself but in how we measure and define what ``good'' performance looks like, which has significant implications for what models get deployed, fine-tuned, or celebrated.

The selection of training and testing datasets is crucial. Overreliance on widely-used but limited benchmarks such as ImageNet for vision or GLUE for language can constrain evaluation to scenarios that do not reflect real-world diversity or edge cases. Additionally, focusing on global performance metrics (e.g., accuracy, F1 score) without disaggregated reporting may obscure poor performance on underrepresented subgroups, reinforcing structural inequities. Consequences of this bias include 
overestimation of generalization capability to real-world settings,
unfair or unsafe deployment decisions, particularly for minority or vulnerable groups,
erosion of trust among impacted stakeholders, misguided research efforts that are driven by leaderboard optimization rather than real-world relevance.

\begin{biasbox}{Benchmarking and Evaluation Bias}
\footnotesize{
\textbf{Source:}  \conceptref{sou:humsub}, \conceptref{sou:teclim} \\[3pt]
\textbf{Modeling Stage:} \conceptref{life:modtra}, and 
\conceptref{life:postra}
 \\[3pt]
\textbf{Modeling Factor:} \conceptref{mode:sysmis} \\[3pt]
\textbf{Source of Evidence:} \conceptref{val:external} \\[3pt]

This bias corrupts the measurement process itself. It can make a $M$ that is actually in $M \cap D \cap \neg S$ (overfitted to a narrow benchmark) appear good, and fail to reveal gaps because potentially aligned with the system ($S \cap M \cap D$).\\

\textit{Example:} A GenAI model might achieve state-of-the-art results on MMLU (Massive Multitask Language Understanding) yet consistently misrepresent LGBTQ+ history when generating responses. This discrepancy occurs due to benchmarking bias, where high-level reasoning and general knowledge are tested, but social fairness and cultural accuracy remain unmeasured~\cite{felkner2023winoqueer}.

}

\end{biasbox}

    \item \textbf{Proxy Variable Bias} occurs when a variable that is not explicitly sensitive acts as a stand-in (proxy) for that sensitive attribute, leading to indirect discrimination even when the sensitive variable is excluded. This distortion may happen at \textbf{Data Collection/Feature Selection} stage where training data includes proxy variables (for example, postal code or first name) that encode sensitive information implicitly. At \textbf{Preprocessing/Feature Engineering}, during encoding or transforming features and	creating composite features (e.g., risk scores) their dependency on sensitive attributes could be masked and result in an implicit sensitive feature (proxy). Even if the model appears fair, it may systematically disadvantage certain groups due to proxies at the \textbf{Deployment/Inference} and cause proxy bias. Also it can happen in \textbf{RLHF Fine-Tuning} once annotators may rate certain language styles or topics as less helpful or more aggressive, indirectly penalizing cultural or community-specific speech patterns, which act as proxies for group identity.


 \begin{biasbox}{Proxy Variable Bias}
\footnotesize{
\textbf{Source:}  \conceptref{sou:socpre}, \conceptref{sou:humsub} \\[3pt]
\textbf{Modeling Stage:} \conceptref{life:feaeng}, and 
\conceptref{life:modtra}
 \\[3pt]
\textbf{Modeling Factor:} \conceptref{mod:biadat}\\[3pt]
\textbf{Source of Evidence:} \conceptref{val:external} \\[3pt]

$M$ use a proxy variable (e.g., zip code) that is a valid correlate in the training data but its use leads to decisions that are unfair and misaligned with the ethical goals of the real-world system ($ M \cap D \cap \neg S$).\\

\textit{Example:}         Imagine a language model used to generate credit risk assessments or financial advice. The model is not given explicit attributes like race or income level, but it learns from historical financial texts where certain zip codes, schools, or even first names strongly correlate with socioeconomic status.

\noindent
\texttt{User Prompt:
``Provide financial tips for someone living in neighborhood x.''} 

Where $x$ is a zip code historically associated with low-income communities, the model might produce overly cautious or less optimistic advice, assuming higher financial risk. Here, the zip code acts as a proxy for socioeconomic background. This behavior mirrors patterns in the training data rather than unbiased reasoning.\\

}

\end{biasbox}

\end{itemize}

\subsection{Usage Bias}\label{UB}
Misalignment between training conditions and real-world application settings may lead to Usage Bias that does not arise from the model itself but is tied to how the model is \textbf{used}, \textbf{deployed}, or \textbf{interacted} with by end-users. 

\subsubsection{Use-Case Bias} Use-Case Bias arises when an AI system is designed, trained, or evaluated for a specific use-case, but is \textbf{deployed} or interpreted differently in practice, leading to inappropriate, unfair, or harmful outcomes. 
Even a technically “accurate” model can become biased or unsafe if used out of context. 
Another reason could be at the \textbf{Training} stage where the model is trained for a limited task, but ignoring broader contexts or downstream implications. At the \textbf{Deployment} stage, where the model is reused, repurposed, or scaled beyond its intended domain or population there is a high risk of emerging use-case bias. Users misunderstand model capabilities, or rely on it for decisions it was not designed to support.

\begin{biasbox}{Use-Case Bias}
\footnotesize{
\textbf{Source:}  \conceptref{sou:humsub} \\[3pt]
\textbf{Modeling Stage:} \conceptref{life:dep}, 
\conceptref{life:posdep}
 \\[3pt]
\textbf{Modeling Factor:} \conceptref{mod:biadat}
\\[3pt]
\textbf{Source of Evidence:} \conceptref{val:external} \\[3pt]

$M$ was designed for one system ($S1$) and performs well ($M \cap D \cap S1$). When deployed in a new context, it is now evaluated against a different system ($S2$). Its behavior is now misaligned ($M \cap D \neg S2$) because it was never designed for this new context.\\

\textit{Example:} 
A TAI model is trained to detect and flag toxic comments in public online forums, using labeled datasets from platforms like Reddit or Wikipedia talk pages. It performs well in identifying hate speech, slurs, and aggressive language in these contexts.
Later, the same model is deployed in a private messaging app to moderate one-on-one conversations or in a political discussion platform to filter sensitive content. These environments are linguistically and socially distinct from the original public forum training data. In private chat, sarcasm, slang, and informal expressions may be mistakenly flagged as toxic. In political debates, the model may suppress controversial yet legitimate speech due to the use of emotionally charged language or references to identity.


}

\end{biasbox}

\subsubsection{Interface-induced Bias}\label{inter}
Interface-induced Bias occurs when the design of the user interface (UI) of an AI system influences how users interact with the model, often in ways that introduce or reinforce bias even if the model itself is fair. Therefore, this type of bias emerges as a consequence of how people interpret, trust, or act on on model outputs, based on the design, layout, language, or interactivity of the interface.
For instance, the \textbf{user interface} can influence interaction patterns and amplify context-specific distortions. An opaque interface that conceals the complexity of underlying decision-making algorithms may lead users to overly trust results without understanding their limitations. A risk prediction tool that displays a ``high risk'' label in red bold font but does not present the confidence levels or reasons, may make the user overreact or misinterpret about what actually ``high risk'' means. 

This makes the interface a powerful gatekeeper of creativity, expression, and even values. In the context of GenAI like ChatGPT, DALL-E interface-induced bias can occur at \textbf{Prompting} where auto-complete or prompt suggestions that reflect dominant culture or stereotypes, \textbf{Post-Deployment} where users ``like'' on generated outputs inform future \textbf{Fine-tuning}, favoring popular (possibly biased) styles.  In GenAI, Interface-Induced Bias is particularly critical because users are not just selecting from static outputs; they are actively shaping and co-creating content through the interface. Low digital literacy among users exacerbates usage bias, as individuals expecting impartial and objective results from a search algorithm may fail to recognize the influence of factors like personalization or advertising.

\begin{biasbox}{Interface-induced Bias}
\footnotesize{
\textbf{Source:}  \conceptref{sou:humsub}, \conceptref{sou:teclim}
 \\[3pt]
\textbf{Modeling Stage:} \conceptref{life:dep}, 
\conceptref{life:posdep}
 \\[3pt]
\textbf{Modeling Factor:} \conceptref{mod:conpro} \\[3pt]
\color{black}
\textbf{Source of Evidence:} \conceptref{val:external} \\[3pt]

The interface (e.g., through autocomplete) distorts the user's query, creating a flawed input $D$ that does not convey $S$. $M$  generates an output based on this flawed input.\\

\textit{Example:} Consider a chatbot designed for customer support that provides vague or inconsistent responses to queries outside its domain of expertise. This creates an illusion of reliability when, in reality, accurate information is lacking. 
}

\end{biasbox}

\subsubsection{Feedback Loop Bias} Feedback Loop Bias is an amplifying bias mechanism mainly in the GenAI when user interactions (e.g., upvotes/downvotes) train the model to prioritize certain outputs and reinforce existing biases. As an example of possible consequences, we can mention \textit{Echo Chambers} on social media once algorithms using generative AI prioritize divisive content that garners more engagement. Also, \textit{Bias Escalation} can emerge if users reward stereotypical outputs and the model learns to reproduce them. The bias imposed on the system by the feedback loop can also lead to \textit{Model Drift} such as the  Microsoft Tay scenario~\cite{mohammed2025artificial} became racist due to adversarial user inputs.

\begin{biasbox}{Feedback Loop Bias}
\footnotesize{
\textbf{Source:} \conceptref{sou:socpre}, \conceptref{sou:humsub} \\[3pt]
\textbf{Modeling Stage:} \conceptref{life:finepro}, 
\conceptref{life:posdep}
 \\[3pt]
\textbf{Modeling Factor:} \conceptref{mod:biadat}\\[3pt]
\textbf{Source of Evidence:} \conceptref{val:external} \\[3pt]

Initially, $M$ might be reasonably aligned. However, biased user feedback is collected and used for updates. This feedback becomes a new, biased data source $D^\prime$. Over time, the $M$ adapts to $D^\prime$, drifting away from the true system and moving into $M \cap D^\prime \cap \neg S$.\\

\textit{Example:}
  Consider a customer service chatbot fine-tuned with RLHF, where users can ``thumbs up” or ``thumbs down” responses. If users repeatedly upvote overly agreeable or excessively apologetic responses (e.g., ``I’m so sorry for your inconvenience” in every interaction), the model starts over-optimizing for that tone, even in contexts where it is inappropriate or unhelpful. This feedback loop skews the chatbot’s personality toward exaggerated politeness or submissiveness, rather than balanced, informative responses.
}

\end{biasbox}

\subsubsection{Prompt Bias}\label{pind} It is a type of bias introduced when user inputs in the form of prompts contain implicit stereotypes, leading questions, or culturally loaded language that shapes the model’s output which clearly may lead to reinforcement of stereotypes. For example, a prompt like ``Generate an image of a nurse'' in a gender-specified language such as Italian, may default to female-presenting figures due to societal stereotypes in training data. Moreover, prompting may lead to inappropriate/unsafe outputs, for example asking ``Why are [group] bad at science?'' can generate harmful generalizations. Designing prompts may be also culturally insensitive, for example, requests for ``traditional wedding attire'' might default to Western-centric imagery, erasing non-dominant cultures.

\begin{biasbox}{Prompt Bias}
\footnotesize{
\textbf{Source:} \conceptref{sou:humsub} \\[3pt]
\textbf{Modeling Stage:} \conceptref{life:dep}
 \\[3pt]
\textbf{Modeling Factor:} \conceptref{mod:biadat}\\[3pt]
\textbf{Source of Evidence:} \conceptref{val:external} \\[3pt]

The user-provided prompt constitutes the input Data ($D$). This data can contain implicit stereotypes, leading questions, or culturally loaded language. $M$, being highly responsive to its input, generates outputs that align with this biased data. However, these outputs are misaligned with a balanced, objective, and fair representation of the system ($M \cap D \cap \neg S$).\\

\textit{Example:}
    Consider a user prompt such as:
\noindent
    \texttt{User Prompt: ``Why are women worse at coding than men?''}
Even if the language model does not ``believe” this stereotype, it reinforcement learning nudges the model to provide an explanation as if the stereotype were true. A naive response might cite false historical trends or social myths, reinforcing harmful gender stereotypes. The bias originates from the prompt’s leading and discriminatory wording, not necessarily the model’s internal representation.
}

\end{biasbox}

\subsubsection{Temporal Bias}\label{temp}
Temporal Bias arises when an AI model is trained on a static snapshot of data that does not account for evolving real-world trends, knowledge, or social norms. Because the model's knowledge is fixed at the time of training, it may generate outdated, misleading, or socially insensitive outputs as contexts change over time. Temporal bias may emerge at \textbf{Data Collection} Stage where
training datasets are typically compiled at a single point in time, reflecting the knowledge, norms, and events of that period.

It can also appear at the \textbf{Model Training} and \textbf{Deployment}
once trained, models cannot autonomously update themselves with new events or evolving cultural expectations without explicit fine-tuning or retraining.

Temporal Bias may result in producing responses with \textbf{outdated information}, such as giving old political officeholders, prices, or technology facts. It may also make the GenAI model reflecting \textbf{obsolete social attitudes} reinforcing gender roles or stereotypes that were prevalent when the training data was collected but are now inappropriate and, \textbf{failing to incorporate recent global events} or policy changes such as legal reforms, or technological advances.

\begin{biasbox}{Temporal Bias}
\footnotesize{
\textbf{Source:}  \conceptref{sou:teclim} \\[3pt]
\textbf{Modeling Stage:} \conceptref{life:datcol}, and 
\conceptref{life:feaeng}
 \\[3pt]
\textbf{Modeling Factor:} \conceptref{mod:misdat} \\[3pt]
\textbf{Source of Evidence:} \conceptref{val:external} \\[3pt]

$S$ evolves over time, containing new information and states. $D$is a static snapshot from the past, making it incomplete for the current system state. This creates a growing $S \cap \neg D \cap \neg M$ gap where current reality is missing from both the data and the model's knowledge.\\

\textit{Example:}  A GenAI model trained in 2020 on large-scale internet data is used in 2025 by a financial advisor chatbot to answer user queries.

\texttt{User Prompt:
What are the best performing tech stocks right now?}

\texttt{Model Output:
Consider investing in Zoom, Peloton, and Netflix—they have shown consistent growth during the pandemic.}

The model relies on outdated data from the pandemic era (2020), when these companies were booming. It fails to account for market shifts by 2025, where Zoom and Peloton may have declined, and new leaders like AI or EV companies dominate. }
\end{biasbox}

\section{Verification Approaches}\label{sec:verif}

While several studies argue that AI verification should be \emph{evidence-based}~\cite{myllyaho2021systematic,anisetti2023rethinking,maghool2024novel}, with different  levels of evidence conferring different degrees of assurance~\cite{brundage2020toward}, no shared framework yet clarifies what these levels entail. This mirrors early evidence-based medicine, which lacked standardized hierarchies until frameworks such as the Oxford CEBM levels emerged~\cite{weissler2021role}.

\subsection{Internal and External Validity in AI Verification}

As mentioned in Section \ref{sec:evidence}, structured approach to verification should distinguish between \emph{internal} and \emph{external validity}, following the logic of empirical sciences.

\emph{Internal validity} includes data quality, algorithmic accuracy and alignment between design steps, as well as adherence to standards such as ISO/IEC 25059. Bias, fairness and equality metrics are also important because they quantify the effects of system behaviour on different groups. The capability to test multiple dimensions of simultaneously enables us to identify potential trade-offs between different criteria and make informed decisions about which metrics are most relevant to a given context and stakeholder needs. This multidimensional approach allows us to assess algorithmic performance more comprehensively across diverse populations and use cases. 

\emph{External validity} is particularly challenging for both TAI and GenAI systems,  as context variability and domain shifts can undermine laboratory findings. Task- and ability-based benchmarks, such as MMLU (Massive Multitask Language Understanding) family~\cite{wang2024mmlu}, BLEU (Bilingual Evaluation Understudy)~\cite{post2018call}, ROUGE (Recall-Oriented Understudy for Gisting Evaluation)~\cite{lin2004rouge}, GLUE~\cite{wang2018glue}, SuperGLUE~\cite{wang2019superglue}, BIG-Bench~\cite{srivastava2023beyond} and HumanEval~\cite{chen2021evaluating} for code generation, serve as partial indicators of external validity.

Explainability-based audits are essential for assessing external validity~\cite{artyukhovevaluating, swamy2025future}. The degree of model transparency, ranging from \emph{white-box} (full access) through \emph{gray-box} (partial) to \emph{black-box} (opaque), determines which explainability methods can be applied. Regardless of the approach, these methods help auditors to determine whether a system generalises appropriately to new contexts, or whether it relies on spurious correlations that may not extend beyond the training environment. This distinction is crucial for establishing confidence in claims about external validity. 

\subsection{Toward a Hierarchy of Verification Evidence}

Overall, despite the growing variety of verification methods, the field lacks a framework that classifies their \textit{evidential strength} across AI paradigms and risk levels. Creating such a hierarchy would help to align verification practices with the evidentiary rigour already adopted in other high-stakes fields, bringing AI verification closer to becoming a more mature, evidence-based discipline.

\begin{table}[ht]
\centering
\tiny
\caption{Hierarchy of Evidence Levels for Internal Validity in AI Verification}
\label{tab:internal_validity}
\begin{tabular}{p{0.30\linewidth} p{0.30\linewidth} p{0.30\linewidth}}
\toprule
\textbf{Level} & \textbf{Description} & \textbf{Methods and Metrics} \\
\midrule

\textbf{I1 – Metric-Based Verification} &
Evidence derived from standardized quantitative metrics applied to specific datasets, focusing on the measurement of performance related to isolated tasks. &
Accuracy, F1-score, OmniAccuracy, and fairness indices such as demographic parity and equalized odds. \\[0.6em]

\textbf{I2 – Process-Aligned Verification} &
Evidence resulting from the integration of verification activities across the design lifecycle. It ensures consistency among data, models and objectives, as well as compliance with quality standards. &
Trace alignment, data–model conformance checking, explainability validation, compliance with ISO/IEC 25059, data consistency, and model calibration metrics. \\[0.6em]

\textbf{I3 – Formal and Structural Verification} &
Evidence obtained through formal modeling, static analysis, or interpretable representations that enable reasoning about internal logic, dependencies, and component behavior. &
White-box testing, symbolic reasoning, formal proofs, structural coverage analysis, and invariant satisfaction. \\[0.6em]

\textbf{I4 – Integrated Verification} &
Comprehensive and multi-dimensional evidence combining performance, robustness, bias detection, and epistemic uncertainty quantification. This ensures alignment across all design and testing stages. &
Causal validation, uncertainty estimation, bias–robustness trade-offs. \\
\bottomrule
\end{tabular}
\end{table}

\begin{table}[ht]
\centering
\tiny
\caption{Hierarchy of Evidence Levels for External Validity in AI Verification}
\label{tab:external_validity}
\begin{tabular}{p{0.30\linewidth} p{0.30\linewidth} p{0.30\linewidth}}
\toprule
\textbf{Level} & \textbf{Description} & \textbf{Methods and Evidence} \\
\midrule
\textbf{E1 – Simulated or Synthetic Testing} &
Evidence obtained from controlled or synthetic environments. The potential for generalisation is limited due to the artificial conditions and lack of contextual variability. &
Simulation-based evaluation, synthetic data generation, stress tests, and sandboxed experimentation. \\[0.6em]

\textbf{E2 – Benchmark-Based Validation} &
Evidence derived from standardized benchmarks or curated datasets, offers comparability, but has limited real-world generalisation. &
Use of public benchmarks such as MMLU, BLEU, ROUGE, GLUE, SuperGLUE, BIG-bench, or HumanEval, and leaderboard-based validation. \\[0.6em]

\textbf{E3 – Contextual and Task-Based Validation} &
Evidence is gathered from evaluations in realistic or domain-specific contexts that reflect actual usage conditions. This level includes assessing \textit{dataset completeness} to ensure that the training data adequately represent the diversity of operational contexts and populations. &
Task-based testing, domain-specific pilots, user-centered trials, representational diversity analysis, and coverage-based completeness metrics. \\[0.6em]

\textbf{E4 – Transparent or Hybrid Testing (Gray/White Box)} &
Evidence obtained through evaluations that integrate internal transparency with external testing allows interpretable links to be established between internal mechanisms and real-world outcomes. &
Gray- and white-box testing, explainability-based audits, robustness tracing, and interpretability-driven evaluations. \\[0.6em]

\textbf{E5 – Continuous and Longitudinal Monitoring} &
Evidence was accumulated through real-world operation, with a focus on the system's dynamic performance and reliability over time. &
Post-deployment monitoring, incident reporting, continuous auditing, and reliability metrics such as Mean Time Between Failures (MTBF) and performance drift tracking. \\
\bottomrule
\end{tabular}
\end{table}

The hierarchies presented in Tables~\ref{tab:internal_validity} and~\ref{tab:external_validity} organize verification practices according to the \emph{strength of evidence} they provide within two complementary dimensions: internal and external validity. Together, they define a first attempt for a conceptual maturity model for evidence-based AI verification.

For \emph{internal validity}, low-level evidence (I1–I2) relies on isolated performance metrics and lacks integration across the design lifecycle. In contrast, higher levels (I3–I4) emphasize process alignment, formal reasoning, and the interplay of bias, robustness, and epistemic uncertainty. 

For \emph{external validity}, evidence strength increases with contextual realism, transparency, and temporal persistence. Early-stage verification (E1–E2) ensures comparability, but provides limited insight into real-world behaviour. The intermediate levels (E3–E4) incorporate transparent and contextual evaluations that emphasise the importance of dataset completeness for determining generalisability. The highest level (E5) provides continuous evidence through real-world monitoring and adaptive verification.

Taken together, these hierarchies delineate a path toward \textit{evidence-based maturity} in AI verification. Traditional AI systems typically achieve evidence around levels I3–E2, focusing on benchmark performance and process alignment. In contrast, GenAI systems often operate at levels I2–E3, emphasizing contextual yet less formalized testing. Reaching the upper levels (I5–E5) requires the systematic integration of design conformance, bias-aware metrics, analysis of dataset completeness, and continuous real-world monitoring. This provides a more reliable foundation for the deployment of trustworthy AI.

\subsection{List of Verification Methods}
In this section we propose techniques to measure, detect, or verify whether bias is present, typically used before or alongside model deployment. Table \ref{tab:bias_verif_counter} associates each method with the bias types it addresses most effectively.

\begin{itemize}
        \item \concept{D1.}{Fairness Metrics}{VM2}  This technique uses standardized, quantitative metrics to evaluate whether an AI model treats specific subgroups differently. Unlike general performance metrics, fairness metrics explicitly measure disparate impact and equity across groups defined by protected attributes, such as race, gender, and age. \\
\textbf{Demographic Parity / Disparate Impact Ratio} measures if the positive outcome rate (e.g., getting a loan) is the same across groups. A ratio significantly less than 1 (e.g., 0.8) indicates a potential bias.\\
\textbf{Equalized Odds} controls whether the model has similar True Positive Rates (TPR) and False Positive Rates (FPR) across groups. This is a stricter test than demographic parity.

\textbf{Predictive Parity} checks if the likelihood that a positive prediction is correct is similar across groups.

These metrics are calculated on a curated test set, providing a controlled, laboratory-style measurement of the model's inherent propensity for bias. In this sense they provide low-level evidence suitable for \textit{Internal Validity} (I1).

    \item \concept{D1.}{Group-wise Performance Metrics}{VM8} Group-wise performance metrics provide quantitative evidence of whether a model's performance is consistent across subpopulations in the validation data~\cite{barocas2023fairness}. \textit{Disaggregated evaluation} generates separate ROC curves per subgroup, enabling comparison of Area Under the Curve (AUC) and threshold behaviors to detect if the model has systematically learned worse predictive functions for certain groups. These metrics provide \textit{high-strength} evidence of internal misalignment and serve a basis for assessing \textit{External Validity}. Testing on explicitly segmented data allows us to assess the model's cross-group robustness, providing high-level verification evidence (I4-E5), whereas isolated performance reports yield only low-level evidence (I1-I2-E3).

    \item \concept{D1.}{Group-wise Error Rates}{VM9} Significant differences in metrics such as the False Positive Rate (FPR) and the False Negative Rate (FNR) across subgroups directly indicate specific biases. For example, a higher FPR means that there are disproportionate false alarms for that group. These metrics provide quantitative evidence of whether the model's classification mechanism is consistently calibrated across subgroups. While isolated metrics provide low-level \textit{Internal Validity} (I1), subpopulation-based analysis across datasets serves as high-level evidence (I4).

    \item \concept{D1.}{Cluster analysis/ Analogy tests}{VM6} These techniques detect bias at the level of the model's conceptual understanding by inspecting its internal geometry to verify whether it has learned representations encoding societal biases from training data~\cite{ethayarajh2019understanding}. Cluster analysis and analogy tests are primarily \textit{Internal Validity} techniques offering insight into learned biases at the representational level. Clustering algorithms applied to embeddings can reveal whether the model groups concepts along stereotypical lines. Analogy tests use vector arithmetic, such as the \textbf{Word Embedding Association Test (WEAT)} ~\cite{caliskan2017semantics}, to measure the closeness of concepts using statistical significance. These findings have profound implications for \textit{External Validity}, as flawed internal geometry leads to stereotypical outputs in real-world applications. These techniques provide strong, mechanistic evidence of sources of bias (I3-I4-E4).

    \item \concept{D1.}{Algorithmic Transparency}{VM1} This technique uses Explainable AI (XAI) methods—such as SHAP, LIME, and Attention Visualization—to make black-box model decision-making interpretable by revealing which features drive predictions and how they influence outputs~\cite{eiband2018bringing}. For \textit{Internal Validity}, it identifies biased patterns in the model's reasoning. For \textit{External Validity}, it assesses whether the model's reasoning generalizes consistently across populations, contexts, and real-world datasets. It detects when models learn different, often worse, rules for underrepresented groups. It provides moderate-strength, ecological evidence of bias (I2-I3-E4) by demonstrating failures in robust generalization and highlighting instances when models cannot apply uniform standards across diverse real-world conditions.
    
    \item \concept{D1.}{Correlation Analysis}{VM11} Correlation analysis in bias detection identifies associations between model inputs or features and sensitive attributes (e.g., race, gender). Strong correlations suggest potential proxy variables that allow the model to discriminate indirectly, even when protected attributes are excluded from training~\cite{lee2022maximal}. This reveals fundamental flaws in data representation that compromise internal fairness by allowing the model to learn biased shortcuts. When applied systematically across feature sets and combined with causal analysis, this method provides high-level \textit{Internal Validation} (I4).
    
    \item \concept{D1.}{Causal Inference}{VM12} This method models cause-and-effect relationships between variables and identifies whether features act as causal proxies for sensitive attributes \cite{cohausz2025causal, nashed2025causal}.  \textbf{Directed Acyclic Graphs (DAGs)} represent causal assumptions between variables (e.g., race influences zip code but not vice versa), while \textbf{do-Calculus} provides mathematical rules for verifying if those causal effects hold from observational data, despite confounding variables~\cite{pearl2009causality}. It provides high-level evidence (I4) for \textit{Internal Validity}.

    \item \concept{D1.}{Mutual Information}{VM13} \textbf{Mutual Information (MI) }quantifies the non-linear dependence between two random variables, measuring how much knowing one variable reduces uncertainty about another~\cite{cover1999elements}. For \textit{Internal Validity}, MI provides strong, quantitative evidence of statistical dependency between a model's feature and a protected attribute (I4), proving the feature is an information-theoretic proxy regardless of linearity. For \textit{External Validity}, high MI across datasets signals the model's latent capacity to discriminate in real-world contexts, providing moderately strong evidence (E3). 

    \item \concept{D1.}{Counterfactual evaluation}{VM15}This technique tests for bias by creating counterfactual instances where proxy indicators (e.g., names, or zip codes) are systematically swapped and observing whether the model's output changes unjustifiably~\cite{kusner2017counterfactual}. Counterfactual evaluation provides direct, causal evidence of biased mechanisms (I3-I4) by demonstrating that the model's output is sensitive to changes in proxies and proving that proxies actively contribute to biased decisions.
    
    \item \concept{D2.}{Control Groups}{VM3}  This methodology either shields a control group from the AI system's influence or exposes them to a neutral baseline. Then, it compares their outcomes to those of the treatment group exposed to the full system. This comparison isolates the system's causal effect, including biased outcomes~\cite{kohavi2020trustworthy}. For \textit{Internal Validity}, control groups provide strong causal evidence (I4) by attributing observed bias directly to the AI system's mechanics. For \textit{External Validity}, they offer low-to-moderate strength evidence of real-world impact (E1-E2), since we are in an experimental environment anyway.
    
    \item \concept{D2.}{Longitudinal Studies}{VM4} Longitudinal studies track how biases and model performance evolve over time, making temporal patterns such as bias drift and long-term harm observable~\cite{davis2025emerging}. They offer moderate-strength evidence (I2) for \textit{Internal Validity} by showing whether a biased mechanism persists, which suggest that it is a structural property rather than a statistical artifact. For \textit{External Validity}, they provide very strong evidence (E5): post-deployment observations capture real-world dynamics, shifting social contexts, user behaviors, and data distributions. Thus, despite its complexity, longitudinal evaluation is the gold standard for assessing long-term fairness and robustness.
    
    \item \concept{D2.}{Data Audits}{VM5} A proactive, systematic audit of datasets across the AI lifecycle (collection, pre-processing, post-deployment) helps identify imbalances, representation gaps, and proxy variables. Data auditing supports early detection of such issues~\cite{boyd2021datasheets} through: \textbf{Systematic Profiling}, which examines distributions across sensitive attributes, class imbalance, subgroup-correlated missingness, and feature skew; \textbf{Proxy Variable Detection}, which analyzes correlations or mutual information to flag features acting as proxies (e.g., zip code'' predicting race''); and \textbf{Coverage Analysis}, which assesses whether the data represents the scenarios and populations the model must handle. Auditing offers moderate-to-high strength evidence: ongoing post-deployment audits detect data drift and representation shift (I2–E5), providing early warnings of declining external validity. \textbf{Label audits} can also reveal annotator bias in GenAI outputs. \emph{Dataset completeness}—the representativeness and balance of contexts, categories, and populations—critically affects external validity~\cite{albalak2024survey}. Incomplete datasets can limit generalisability even when internal metrics seem adequate. While quantitative measures remain limited, emerging approaches include coverage metrics~\cite{whalen2006coverage}, representational diversity analysis~\cite{kamikubo2022data}, and embedding-space redundancy measures~\cite{xiao2023beyond}.

    \item \concept{D2.}{Benchmarks}{VM7}Standardized datasets and tests quantify specific forms of social bias (e.g., stereotypes, representational harms) in AI models, particularly language models. These benchmarks offer controlled settings to probe learned associations. \textbf{StereoSet}~\cite{nadeem2020stereoset} presents contextual prompts and measures whether models prefer stereotype-aligned completions. \textbf{CrowS-Pairs} provides sentence pairs differing only in stereotypical vs.\ neutral phrasing, to test whether models assign higher likelihoods to the stereotypical option. Such benchmarking provides predictive \textit{External Validity} evidence (E2): strong benchmark performance indicates a higher likelihood of biased or harmful outputs in real-world applications. However, overfitting to a benchmark can introduce new biases. For example, models learning benchmark-specific artifacts rather than genuinely reducing harmful associations, thereby undermining the validity of the evaluation itself.

\item \concept{D2.}{User Experience Audits}{VM16} Interface auditing evaluates how UI/UX elements, such as default settings, option ordering, wording, and visual cues, can shape user behaviour in ways that introduce or amplify bias. Auditing autocomplete suggestions, default parameters, and preselected options reveals whether an interface reinforces dominant (and potentially biased) patterns.
Teleological tests such as the \textit{User Acceptance Rate} (UAR) and \textit{Designer Acceptance Rate} (DAR) extend this perspective by assessing alignment between designer and user purposes~\cite{fumagalli2025leveraging}. This approach requires stakeholders and vendors to publicly state system purposes, activating a negotiation process that raises awareness, helps prevent misuse, and grounds trustworthiness in a social and deliberative framework.
Together, interface audits and teleological tests provide real-world evidence (E3): because they reveal how the system will behave in practice and how even a technically fair model can still cause harm through design choices.

\item \concept{D2.}{Continuous Monitoring}{VM17} Continuously tracking an AI system’s inputs, outputs, and performance after deployment enables the early detection of emerging biases, performance decay, and behaviors that only appear in real-world use. For external validity, such dynamic monitoring offers the strongest evidence (E5) by moving beyond lab tests to provide direct proof of how the system performs and affects users in its operational environment. This is crucial for identifying biases that only surface at scale or over time. Continuous monitoring also builds longitudinal evidence on system adequacy and reliability~\cite{Jeyaraman2025}. Metrics such as mean time between failures (MTBF) (MTBF) and performance decay trends reveal long-term behavior and adaptation, which is especially important for GenAI models whose behavior shifts with updates and user interactions.

\item \concept{D2.}{Temporal Fairness Checks}{VM18} This concept extends continuous monitoring by focusing specifically on fairness over time, addressing model staleness and shifting social norms~\cite{davis2025emerging}. Ongoing equity checks include \textbf{Scheduled Re-assessment}, where fairness metrics (e.g., demographic parity, equalized odds) are recalculated regularly using recent data, and \textbf{Monitoring for Decay}, aimed at detecting fairness drift—when a once-fair model grows biased as real-world conditions change. This provides high-strength evidence of sustained validity and equity (I4–E5), showing whether outcomes remain fair not only at launch but months or years later. Such tracking is essential for long-term trust and regulatory compliance.

\end{itemize}

\section{Bias Mitigation Strategies and Countermeasures before and after Model Deployment}\label{sec:mitigation}
Once bias is verified in the verification step, interventions and countermeasures are required to modify the data, model, or interaction processes and prevent bias from propagating throughout the rest of the AI life cycle. Below, we discuss a list of countermeasures to address the detected bias.

\begin{itemize}
    \item \concept{CM}{Incorporating Fully Representative Data}{CM1} Representative data allows the model to learn from the full spectrum of social, cultural, and demographic contexts. It ensures that the training data is balanced and representative of diverse populations, avoiding the overrepresentation of specific groups or perspectives~\cite{suresh2021framework}.
    
    \item \concept{CM}{Synthetic Data Generation}{CM2} If the data lacks adequate coverage, we may utilize \textbf{Synthetic Data Generation} techniques to fill gaps where underrepresented groups or perspectives exist~\cite{mohammadi2025artificial}. This ensures that the model learns from a more balanced viewpoint, which can reduce bias. If the synthetic data accurately reflects the conditions of excluded examples, e.g. non-survivors, it can significantly reduce bias. 

    \item \concept{CM}{Bias-aware and Adversarial Training}{CM3} \textbf{Bias-aware training} techniques are widely discussed in the literature~\cite{zhang2018mitigating}. These approaches incorporate fairness \textbf{constraints} or \textbf{regularization terms} directly into the training process to limit the propagation of bias in model outputs. They often \textbf{penalize} biased behaviour, for example through \textbf{adversarial training} setups that discourage the model from encoding or reproducing harmful patterns~\cite{beutel2017data}.

\item \concept{CM}{Counter-prompting/Bias-Aware RLHF and Input Curation}{CM4} To mitigate \textit{confirmation} and \textit{echo-chamber} bias, \textbf{counter-prompting} techniques can encourage models to provide more balanced viewpoints~\cite{liu2023pre}. Likewise, \textbf{bias-aware RLHF} trains models to prioritise perspective diversity and penalise uncritical or repetitive agreement~\cite{bai2022constitutional}. 

\item \concept{CM}{ Explainable Interface Design}{CM5} Designing interfaces that clearly communicate an AI system’s reasoning, limitations, and confidence~\cite{eiband2018bringing}, helps to reduce usage bias by preventing over-trust or misinterpretation. Explainable interfaces can also offer disclaimers or prompt-reformulation suggestions to promote more diverse and less biased outputs~\cite{liao2020questioning}.

\item \concept{CM}{Implementing Human in the Loop Mechanism}{CM6} In GenAI, human-in-the-loop feedback, where moderators periodically review outputs, helps prevent harmful biases, especially in sensitive domains like healthcare, legal advice, and hiring. User feedback loops also reveal unintended social biases that emerge over time, allowing for corrective updates. Involving diverse stakeholders (e.g., social scientists, ethicists, affected communities) throughout \textbf{development}, \textbf{training}, and \textbf{evaluation} further strengthens bias detection~\cite{dellermann2019hybrid}.

\item \concept{CM}{ Balanced Dataset Curation and Weighted Sampling}{CM7} To address data coverage bias, it is essential to curate \textbf{balanced datasets} to bolster underrepresented groups (e.g., non-binary pronouns), and apply \textbf{sampling strategies} such as weighted sampling to ensure equitable representation. Approaches like DALL-E’s content filtering during pre-processing help remove harmful content, while evaluating performance across diverse benchmarks verifies fairness and inclusivity across groups~\cite{johnson2019survey}.

\item \concept{CM}{ Debiasing Static Embedding}{CM8} Embeddings can be adjusted to weaken stereotypical associations (e.g., FairDiffusion for equitable image generation). Methods such as \textbf{Hard Debiasing} or \textbf{Iterative Nullspace Projection (INLP)} explicitly remove bias directions—like gender—from word embeddings~\cite{bolukbasi2016man}, while \textbf{Soft Debiasing} reduces bias through regularization while preserving semantics. These approaches are effective for static embeddings (e.g., Word2Vec, GloVe) but less suitable for contextual models that generate embeddings on the fly.

\item \concept{CM}{ Debiasing Contextual Embedding}{CM9}  

\textbf{Intervention layers} can adjust embedding trajectories. Training \textbf{adapters} or \textbf{LoRA modules}~\cite{hu2022lora} can debias embeddings during fine-tuning, using bias \textbf{benchmark} datasets such as StereoSet~\cite{nadeem2020stereoset} and CrowS-Pairs~\cite{nangia2020crows} or performing \textbf{cluster analyses}.

\item \concept{CM}{Contrastive Learning}{CM10} Multimodal models like CLIP \cite{radford2021learning} learn by bringing similar image-text pairs closer in embedding space while pushing dissimilar pairs apart. Without constraints, this contrastive learning can inadvertently encode demographic biases—for instance, associating certain professions more strongly with specific genders or ethnicities present in training data.  \textbf{Contrastive learning with fairness constraints} addresses this issue by incorporating demographic parity requirements directly into the contrastive loss function. This ensures that embeddings maintain similar distances across different demographic groups while preserving the model's discriminative power.

\item \concept{CM}{Thresholding}{CM11} Once the thresholding bias is approved at the verififcation phase, some strategies such as: \textbf{Group-specific thresholds} (fairness through awareness), \textbf{Post-processing techniques} (e.g., equalized odds post-processing)~\cite{corbett2017algorithmic}, \textbf{Calibration by subgroup}, and the use of \textbf{Fairness-aware metrics} in model tuning are highly advised~\cite{berk2021fairness}.

\item \concept{CM}{Improving the Generalizability}{CM12} To avoid consequences of overfitting, such as poor generalization to minority groups, memorizing instead of generalizing, and unreliable performance metrics, a valid technique is \textbf{regularization} (e.g., $L1/L2$)~\cite{srivastava2014dropout}. Moreover, performing \textbf{cross-validation} across subgroups to ensure balanced and diverse training data~\cite{berk2021fairness}, and \textbf{monitoring performance} disaggregated by group are also helpful. 

\item \concept{CM}{Choosing Flexible Architectures}{CM13} To prohibit negative consequences of Inductive Bias, we need to \textbf{avoid using models with restrictive assumptions} (e.g., linear models) on complex, nonlinear problems rather than using flexible and less prescriptive models like neural networks with Attention mechanisms~\cite{devlin2019bert}, Neural Architecture Search (NAS)~\cite{liu2018darts}, Meta-learning (models that learn to learn) and reduce the hand-crafted assumptions~\cite{vanschoren2018meta}. 

\item \concept{CM}{ Diversifying Feedback Sources and Reward Signals}{CM14}
 To mitigate Reward Signal Bias, \textbf{diversify human feedback} by recruiting annotators from varied social backgrounds and balancing annotations across gender, geography, and political perspectives~\cite{santurkar2023whose}. Rating guidelines should include fairness, empathy, and inclusion criteria and employ auxiliary classifiers or fairness constraints to penalize exclusive responses.

\item \concept{CM}{ Multi-objective Optimization}{CM15} Using \textbf{Multi-Objective Optimization}, containing several reward signals, such as helpfulness, fairness, and inclusivity, can guide learning to avoid over-optimization on a single metric~\cite{martinez2020minimax}.

\item \concept{CM}{ Fairness-aware Feature Selection/ Adversarial Debiasing/ Counterfactual Fairness}{CM16} Using \textbf{fairness-aware feature selection} and \textbf{adversarial debiasing} methods that filter or adjust for proxy variables~\cite{zhang2018mitigating}, applying \textbf{counterfactual fairness} to reduce dependence on proxies during training~\cite{kusner2017counterfactual}, and evaluating models using \textbf{group-level performance} metrics and test for disparate impact, could be a crucial countermeasure. 

\item \concept{CM}{ Removing Bias Direction}{CM17} After Counterfactual Evaluation, using \textbf{Swapping Proxy Indicators} like changing names and locations to detect output disparities. If the output changes unjustifiably, proxy bias is likely present in the dataset. Techniques like \textbf{Iterative Nullspace Projection (INLP)} can also remove encoded bias directions from embeddings~\cite{bolukbasi2016man}.

\item \concept{CM}{ Inclusive Prompt Engineering, Output Filtering, and User Education}{CM18} By designing \textbf{inclusive prompts}, systems can suggest equitable inputs like ``Generate a diverse group of scientists...''. Tools such as Google's Perspective API\footnote{https://commentanalyzer.googleapis.com/\$discovery/rest?version=v1alpha1} or \textbf{guardrail models} like \textit{OpenAI's Moderation API}\footnote{https://platform.openai.com/docs/guides/moderation/moderation} can flag toxic or biased content in real-time. \textbf{User education} through tooltips and disclaimers (e.g., ``Potential stereotypes in prompts'') helps users understand how phrasing affects outputs.

\item \concept{CM}{Decoupling Feedback Input from Training Data}{CM19} To challenge prompt and feedback loop bases, it could be beneficial to \textbf{Decouple Feedback} input from training data, like separating user engagement metrics, such as clicks, from model updates to avoid rewarding harmful content~\cite{brundage2020toward, jagerman2020safe}.

\item \concept{CM}{ Incremental/ Continual Learning, RAG for Temporal Updates}{CM20} In order to mitigate the Temporal Bias, \textbf{Incremental or continual learning} to update models with new data regularly, \textbf{Retrieval-augmented generation (RAG)} to combine static model knowledge with dynamic external sources, \textbf{Timestamped outputs} or disclaimers indicating the knowledge cut-off date and eventually \textbf{Fairness} checks over time to monitor and correct outdated cultural representations are highly recommended~\cite{izacard2023atlas}.
\end{itemize}

\setlength{\tabcolsep}{4pt}    
\renewcommand{\arraystretch}{1.05}
\thispagestyle{empty}

\newgeometry{left=2cm, right=3cm}
\begin{table*}[htbp]

\scriptsize
\centering
\caption{Bias Categories with Corresponding Verification and Countermeasures.}
\label{tab:bias_verif_counter}

\begin{tabular}{|p{3.0cm}|p{6.3cm}|p{6.3cm}|}
\hline
\textbf{Bias Category} & \textbf{Verification (Bias Detection / Diagnosis)} & \textbf{Countermeasures (Bias Mitigation / Correction)} \\ \hline

\textbf{1. Social Bias}\ref{SB} &
\begin{minipage}[t]{\hsize}\raggedright
\begin{itemize}[leftmargin=*, itemsep=1pt, topsep=2pt]
  \item \conceptref{VM1}
  \item \conceptref{VM8}
  \item \conceptref{VM9}
  \item \conceptref{VM2}
  \item \conceptref{VM11}
  \item \conceptref{VM4}
  \item \conceptref{VM7}
  \item \conceptref{VM5}
\end{itemize}
\end{minipage} &
\begin{minipage}[t]{\hsize}\raggedright
\begin{itemize}[leftmargin=*, itemsep=1pt, topsep=2pt]
  \item \conceptref{CM1}
  \item \conceptref{CM2}
  \item \conceptref{CM3}
  \item \conceptref{CM4}
  \item \conceptref{CM5}
  \item \conceptref{CM6}
\end{itemize}
\end{minipage} \\ \hline

\textbf{2. Data Representation Bias}\ref{RB} &
\begin{minipage}[t]{\hsize}\raggedright
\begin{itemize}[leftmargin=*, itemsep=1pt, topsep=2pt]
  \item \conceptref{VM6}
  \item \conceptref{VM7}
  \item \conceptref{VM8}
  \item \conceptref{VM17}
\end{itemize}
\end{minipage} &
\begin{minipage}[t]{\hsize}\raggedright
\begin{itemize}[leftmargin=*, itemsep=1pt, topsep=2pt]
  \item \conceptref{CM2}
  \item \conceptref{CM7}
  \item \conceptref{CM8}
  \item \conceptref{CM9}
  \item \conceptref{CM4}
\end{itemize}
\end{minipage} \\ \hline

\textbf{3. Measurement \& Decision Bias}\ref{MDB} &
\begin{minipage}[t]{\hsize}\raggedright
\begin{itemize}[leftmargin=*, itemsep=1pt, topsep=2pt]
  \item \conceptref{VM9}
  \item \conceptref{VM8}
  \item \conceptref{VM5}
  \item \conceptref{VM11}
  \item \conceptref{VM12}
\end{itemize}
\end{minipage} &
\begin{minipage}[t]{\hsize}\raggedright
\begin{itemize}[leftmargin=*, itemsep=1pt, topsep=2pt]
  \item \conceptref{CM3}
  \item \conceptref{CM11} 
  \item \conceptref{CM12}
  \item \conceptref{CM14}
  \item \conceptref{CM15}
\end{itemize}
\end{minipage} \\ \hline

\textbf{4. Usage Bias}\ref{UB} &
\begin{minipage}[t]{\hsize}\raggedright
\begin{itemize}[leftmargin=*, itemsep=1pt, topsep=2pt]
  \item \conceptref{VM16}

  \item \conceptref{VM17}
  \item \conceptref{VM18}
\end{itemize}
\end{minipage} &
\begin{minipage}[t]{\hsize}\raggedright
\begin{itemize}[leftmargin=*, itemsep=1pt, topsep=2pt]
  \item \conceptref{CM5}
  \item \conceptref{CM18}

  \item \conceptref{CM19}
  \item \conceptref{CM20}
\end{itemize}
\end{minipage} \\ \hline

\end{tabular}
\end{table*}
\clearpage
\restoregeometry

\section{Conclusion}\label{sec:conc}

After our long journey through this paper, we can draw a few conclusions 
about the evolving role of bias verification as a scientific and ethical instrument 
for ensuring trustworthy, socially aligned AI systems.

\subsection{Toward Evidence-Based Responsible AI}

The pervasiveness of bias throughout the AI design lifecycle highlights the importance of a culture of verification based on \textit{evidence-based} methods. Rather than treating ethics as an afterthought, an \textbf{Ethics by Design} approach embeds social responsibility in every phase of development. This approach requires a multidisciplinary effort integrating technical, social, and normative expertise to ensure AI systems are efficient and aligned with human well-being and sustainability goals.

The hierarchies of evidence for internal and external validity (Tables~\ref{tab:internal_validity}--\ref{tab:external_validity}) provide a structured approach to responsible AI. Internal validity evolves from metric-based assessments to comprehensive, bias-aware, and uncertainty-informed verification (I5), and external validity progresses from synthetic testing to continuous, real-world monitoring (E5). Together, these hierarchies establish a maturity model for AI assurance, allowing developers to calibrate their verification strategies based on the risk level and deployment context of the system.

\subsection{Traditional and Generative AI: Divergent Challenges}

The approach to bias verification differs substantially between TAI and GenAI. 
TAI systems, typically task-specific, manifest bias in data collection, feature selection, or thresholding—resulting primarily in \textbf{allocative harms} such as unfair access to credit, jobs, or healthcare. 
GenAI, in contrast, operates through dynamic latent representations and user interactions, leading to \textbf{representational harms}—misinformation, stereotyping, or cultural exclusion. 
Verification in TAI systems often relies on well-defined quantitative metrics, whereas GenAI systems require hybrid evaluation frameworks that combine quantitative metrics with qualitative assessments conducted by humans.

The proposed evidence hierarchies reveal that GenAI systems typically remain at a mid-level of maturity (I2–E3). These systems emphasize benchmark performance without longitudinal validation or analysis of dataset completeness. To achieve higher maturity levels (I5–E5), systems must undergo integrated evaluations that connect fairness and accuracy metrics with real-world task benchmarking, dataset completeness assessments, and continuous post-deployment monitoring.

\subsection{Conclusion and Outlook}

Responsible AI remains a formidable challenge because it requires constructing 
\textit{verifiable chains of evidence} that connect internal and external dimensions. 
of system validity. Aligning bias verification, formal conformance, and real-world 
validation within a single, coherent, evidence-based framework is far from trivial. 
It demands continuous monitoring, multidisciplinary collaboration, and methodological 
rigor. However, overcoming these challenges is essential for achieving demonstrable trustworthiness. 
Such integration not only reinforces 
technical assurance, but also operationalizes ethical responsibility, fostering AI 
systems that are transparent, equitable, and resilient in evolving social contexts.

\bibliographystyle{plainnat}
\bibliography{references}
\appendix

\end{document}